\documentclass[10pt,a4paper]{article}
\usepackage{amsmath}
\usepackage{amssymb}
\usepackage{graphicx}
\usepackage{subcaption}
\usepackage{float}
\usepackage{hyperref}
\usepackage{booktabs}

\usepackage{chato-notes}

\usepackage[T1]{fontenc}
\usepackage[utf8]{inputenc}

\newcommand{\abs}[1]{\left\lvert#1\right\rvert}

\newcommand{\spcS}{\mathcal{S}}
\newcommand{\siteS}{\mathcal{K}}

\newcommand{\spc}{i}
\newcommand{\ospc}{j}
\newcommand{\site}{k}
\newcommand{\AssMat}{A}
\newcommand{\Assv}{a}
\newcommand{\InterMat}{I}
\newcommand{\EffMat}{Q}
\newcommand{\Effv}{\alpha}
\newcommand{\ResMat}{P}
\newcommand{\Resv}{\rho}
\newcommand{\BioCont}{C}
\newcommand{\BioEff}{z}
\newcommand{\abiov}{x}
\newcommand{\abdv}{y}

\newcommand{\habs}{h}

\newcommand{\evRAI}{$\text{RAI}_{ij}$}

\newcommand{\epsPos}{\epsilon^+}
\newcommand{\epsNeg}{\epsilon^-}

\usepackage{setspace}
\usepackage{tabularx}
\usepackage{enumitem}
\usepackage{natbib}
\usepackage[table,xcdraw]{xcolor}

\usepackage{longtable}

\usepackage{color,url,xcolor,tikz,pgfplots}
\usetikzlibrary{arrows,shapes,positioning,calc,decorations.pathreplacing,pgfplots.statistics}
\usepackage{csvsimple}

\definecolor{darkgrey}{gray}{0.4}
\definecolor{midgrey}{gray}{0.6}
\definecolor{lightgrey}{gray}{0.8}

\definecolor{TolDarkPurple}{HTML}{332288}
\definecolor{TolDarkBlue}{HTML}{6699CC}
\definecolor{TolLightBlue}{HTML}{88CCEE}
\definecolor{TolLightGreen}{HTML}{44AA99}
\definecolor{TolDarkGreen}{HTML}{117733}
\definecolor{TolDarkBrown}{HTML}{999933}
\definecolor{TolLightBrown}{HTML}{DDCC77}
\definecolor{TolDarkRed}{HTML}{661100}
\definecolor{TolLightRed}{HTML}{CC6677}
\definecolor{TolLightPink}{HTML}{AA4466}
\definecolor{TolDarkPink}{HTML}{882255}
\definecolor{TolLightPurple}{HTML}{AA4499}

\definecolor{TolDivA}{HTML}{3D52A1}
\definecolor{TolDivB}{HTML}{3A89C9}
\definecolor{TolDivC}{HTML}{77B7E5}
\definecolor{TolDivD}{HTML}{B4DDF7}
\definecolor{TolDivE}{HTML}{E6F5FE}
\definecolor{TolDivF}{HTML}{FFFAD2}
\definecolor{TolDivG}{HTML}{FFE3AA}
\definecolor{TolDivH}{HTML}{F9BD7E}
\definecolor{TolDivI}{HTML}{ED875E}
\definecolor{TolDivJ}{HTML}{D24D3E}
\definecolor{TolDivK}{HTML}{AE1C3E}

\definecolor{TolRbwA}{HTML}{781C81}
\definecolor{TolRbwB}{HTML}{413B93}
\definecolor{TolRbwC}{HTML}{4065B1}
\definecolor{TolRbwD}{HTML}{488BC2}
\definecolor{TolRbwE}{HTML}{55A1B1}
\definecolor{TolRbwF}{HTML}{63AD99}
\definecolor{TolRbwG}{HTML}{7FB972}
\definecolor{TolRbwH}{HTML}{B5BD4C}
\definecolor{TolRbwI}{HTML}{D9AD3C}
\definecolor{TolRbwJ}{HTML}{E68E34}
\definecolor{TolRbwK}{HTML}{E6642C}
\definecolor{TolRbwL}{HTML}{D92120}

\definecolor{TolSeqA}{HTML}{FFFFE5}
\definecolor{TolSeqB}{HTML}{FFF7BC}
\definecolor{TolSeqC}{HTML}{FEE391}
\definecolor{TolSeqD}{HTML}{FEC44F}
\definecolor{TolSeqE}{HTML}{FB9A29}
\definecolor{TolSeqF}{HTML}{EC7014}
\definecolor{TolSeqG}{HTML}{CC4C02}
\definecolor{TolSeqH}{HTML}{993404}
\definecolor{TolSeqI}{HTML}{662506}

\definecolor{colMut}{HTML}{404096}
\definecolor{colCom}{HTML}{529DB7}
\definecolor{colTro}{HTML}{7DB874} 
\definecolor{colAm}{HTML}{E39C37}
\definecolor{colCmp}{HTML}{D92120}

\tikzstyle{plate circ} = [style={draw, circle, minimum width=2.2em, minimum height=2.2em, inner sep=1, outer sep=1}]
\tikzstyle{plate sqr} = [style={draw, rectangle, minimum width=2.em, minimum height=2.em, inner sep=1, outer sep=1}]
\tikzstyle{plate arrow} = [style={->,shorten <=.5pt,shorten >=.5pt}]
\tikzstyle{pil} = [style={->,thick,shorten <=2pt,shorten >=2pt}]
\tikzstyle{net node} = [style={draw, circle, minimum width=1.8em, minimum height=1.8em, inner sep=1, outer sep=1}]
\tikzstyle{lbl node} = [style={inner sep=2, outer sep=2}]
\tikzstyle{rbl node} = [style={fill=white, inner sep=2, outer sep=2, rounded corners=2}]
\tikzstyle{tbl node} = [style={anchor=base west, align=center}] 
\tikzstyle{pnt node} = [style={fill, circle, inner sep=1}]
\tikzstyle{blg node} = [style={font=\footnotesize, anchor=south}]
\tikzstyle{rll node} = [style={draw, inner sep=2, outer sep=0, rounded corners=2, text width=2.1cm, align=center, minimum height=2.2em, fill=white}]
\tikzstyle{rel node} = [style={font=\tiny, fill=white, opacity=.9, inner sep=.8}]

\tikzstyle{sctSa} = [style={draw, thick, regular polygon, regular polygon sides=3, shape border rotate=-90, minimum height=5pt, text width=2.5pt, inner sep=0.}]
\tikzstyle{sctDa} = [style={fill, regular polygon, regular polygon sides=3, shape border rotate=-90, minimum height=5pt, text width=2.5pt, inner sep=0., opacity=0.8}]
\tikzstyle{sctSs} = [style={draw, thick, regular polygon, regular polygon sides=4, shape border rotate=45, minimum height=3pt, text width=3pt, inner sep=0.}]
\tikzstyle{sctDs} = [style={fill, regular polygon, regular polygon sides=4, shape border rotate=45, minimum height=3pt, text width=3pt, inner sep=0., opacity=0.8}]
\tikzstyle{sctNn} = [style={fill, circle, inner sep=.8, opacity=0.8}]

\colorlet{drP}{TolDivA}
\colorlet{drN}{TolDivK}

\colorlet{cs5P}{drP!40}
\colorlet{cs10P}{drP!70}
\colorlet{cs20P}{drP}

\colorlet{cs5N}{drN!40}
\colorlet{cs10N}{drN!70}
\colorlet{cs20N}{drN}

\pgfplotscreateplotcyclelist{divBlue}{%
{TolDivA, fill=TolDivA, fill opacity=0.8},
{TolDivB, fill=TolDivB, fill opacity=0.8},
{TolDivC, fill=TolDivC, fill opacity=0.8},
{TolDivD, fill=TolDivD, fill opacity=0.8}}

\pgfplotscreateplotcyclelist{divRed}{%
{TolDivK, fill=TolDivK, fill opacity=0.8},
{TolDivJ, fill=TolDivJ, fill opacity=0.8},
{TolDivI, fill=TolDivI, fill opacity=0.8},
{TolDivH, fill=TolDivH, fill opacity=0.8}}

\pgfplotscreateplotcyclelist{divG}{%
  {darkgrey, fill=darkgrey, fill opacity=0.6},
  {darkgrey, fill=darkgrey, fill opacity=0.6},
  {darkgrey, fill=darkgrey, fill opacity=0.6},
  {darkgrey, fill=darkgrey, fill opacity=0.6}}

\pgfplotsset{
	compat=1.8,
	clip = false,
	clip marker paths = true,
	tick align=outside,
    tick align=inside,
    xtick pos=left,
    ytick pos=left,
	x tick label style = {yshift=-1em, anchor=base},
	major tick length = 2pt,
}

\newcolumntype{b}{X}
\newcolumntype{s}{>{\hsize=.6\hsize}X}

\usepackage[margin=1.8cm]{geometry}
\renewenvironment{abstract}
 {\begin{center}
  \bfseries \abstractname
  \end{center}
  \list{}{%
    \setlength{\leftmargin}{2mm}
    \setlength{\rightmargin}{\leftmargin}%
  }%
  \item\relax}
 {\endlist}

\makeatletter
\renewcommand*{\maketitle}{%
\begin{titlepage}
\vfill ~

{\center \Large \bfseries\@title\par}
\smallskip
{\center \large \bfseries\titrun{}\par}

\vfill
{\textsc{Keywords:}
	\keywords{}
}

\vfill

\vspace{5em}
{\textsc{Authors:}
\longauth{}
}

\vfill ~
\end{titlepage}
\newpage
}
\makeatother

\newcommand{\titlong}{Uncovering symmetric and asymmetric species associations from community and environmental data} 

\title{\titlong}
\newcommand{\titrun}{Inferring species association networks [running headline]} 
\author{Sara Si-moussi${}^{1*}$, ${Esther Galbrun}^2$, Giovanni Poggiato${}^1$,${Mickaël Hedde}^3$, ${Wilfried Thuiller}^{1}$}
\newcommand{\longauth}{
\begin{description}[itemsep=1em,topsep=2em]
\item[Sara Si-moussi]~ \url{sara.si-moussi@univ-grenoble-alpes.fr}\\ 
Laboratoire d’Écologie Alpine, CNRS, Univ. Grenoble Alpes, Univ. Savoie Mont Blanc, CNRS, F-38000, Grenoble
\item[Esther Galbrun]~ \url{esther.galbrun@uef.fi}\\
School of Computing, University of Eastern Finland, Kuopio, FI-70211, Finland
\item[Mickaël Hedde]~ \url{mickael.hedde@inrae.fr}\\ 
UMR Eco\&Sols, INRAE, IRD, CIRAD, Montpellier SupAgro, Université Montpellier, F-34000, Montpellier
\item[Giovanni Poggiato]~ \url{giov.poggiato@gmail.com}\\
Laboratoire d’Écologie Alpine, Univ. Grenoble Alpes, CNRS, Univ. Savoie Mont Blanc, CNRS, F-38000, Grenoble\\
\item[Matthias Rohr]~ \url{matthias.rohr@univ-grenoble-alpes.fr}\\
Laboratoire d’Écologie Alpine, Univ. Grenoble Alpes, CNRS, Univ. Savoie Mont Blanc, CNRS, F-38000, Grenoble\\
\item[Wilfried Thuiller]~ \url{wilfried.thuiller@univ-grenoble-alpes.fr}\\
Laboratoire d’Écologie Alpine, CNRS, Univ. Grenoble Alpes, Univ. Savoie Mont Blanc, CNRS, F-38000, Grenoble
\end{description}

\vfill
\noindent
Correspondence to \textbf{Sara Si-Moussi} \\
\textit{Mailing address:} LECA, UMR UGA-USMB-CNRS 5553, Université Grenoble Alpes, CS 40700 38058 Grenoble cedex 9, Francee\\
\textit{E-mail:} \url{sara.si-moussi@univ-grenoble-alpes.fr} 

\vfill
}

\newcommand{\keywords}{network inference, representation learning, probabilistic graphical models, species embeddings, latent variable models, species associations networks}

\begin{document}

\maketitle
\doublespacing

~

\paragraph{Acknowledgements.}
We thank Laura Pollock (McGill University) and Tamara M\"unkenm\"uller (LECA) for guidance on and access to source code for simulating virtual communities; we thank Philippe Choler (LECA) for discussion and crucial explanations on the ecology of Alpine plant communities. We also thank Li Ping Liu for access to source code on exponential family embeddings.

The research was supported by the Agence Nationale pour la Recherche (ANR) through the MIAI@Univ Grenoble Alpes institute (ANR-19-P3IA-0003) and the GlobNet (ANR-16-CE02-0009), Gambas (ANR-18-CE02-0025) and Forbic (ANR-18-MPGA-0004) projects. Most of the computations presented in this paper were performed using the GRICAD infrastructure\!\footnote{
https://gricad.univ-grenoble-alpes.fr}. SMS was initially supported by a joint PhD fellowship between the French National Institute of Agricultural and Environmental Research (INRAE) and the French Research Institute for digital sciences (Inria) and by the Labex Persyval. 


\paragraph{Authorship.}
SS and WT designed the study. SS conceptualized the inference framework with help from EG. WT and SS designed the evaluation methodology. SS, WT and EG analyzed the results. MH gave additional perspectives to the paper. SS and WT wrote the first version of the paper and all authors contributed critically to editing the manuscript.


\paragraph{Source code.}
The source code for running the model is available on this GitHub repository \href{https://github.com/bettasimousss/InferenceEcoAssocNet}{link}.


\vfill

\newpage
~
\vfill

\begin{center}
\begin{minipage}{0.75\textwidth}
\begin{abstract} 
\textbf{Aim} There is no much doubt that biotic interactions shape community assembly and ultimately the spatial co-variations between species. There is a hope that the signal of these biotic interactions can be observed and retrieved by investigating the spatial associations between species while accounting for the direct effects of the environment. By definition, biotic interactions can be both symmetric (e.g. competition, mutualism) and asymmetric (e.g. parasitism, predation, hierarchical competition). Yet, most models that attempt to retrieve species associations from co-occurrence or co-abundance data internally assume symmetric relationships between species. Here, we propose and validate a machine-learning framework able to retrieve  bidirectional associations by analysing species community and environmental data.

\textbf{Innovation} 
Our framework (1) models pairwise species associations as directed influences from a source to a target species, parameterized with two species-specific latent embeddings: the effect of the source species on the community, and the response of the target species to the community; and (2) jointly fits these associations within a multi-species conditional generative model with different modes of interactions between environmental drivers and biotic associations.

Using both simulated and empirical data, we demonstrate the ability of our framework to recover known asymmetric and symmetric associations and highlight the properties of the learned association networks. By comparing our approach to other existing models such as joint species distribution models and probabilistic graphical models, we show its superior capacity at retrieving symmetric and asymmetric  interactions. 

\textbf{Main conclusions} 
Our framework enables ecologists to obtain a more generalized picture of the spatial associations between species without unrealistic assumptions of symmetry. The framework is intuitive, modular and broadly applicable across various taxonomic groups.  

\end{abstract}

\end{minipage}
\end{center}

\vfill

\newpage


\section{Introduction}
Understanding the drivers of species distributions and abundances is a long-lasting goal of biogeography \citep{humboldt1805essai}. Niche theory explains the spatial distribution of species by a set of physiological and adaptive properties allowing them to thrive in specific environmental conditions and decline in others \citep{chase2003ecological,pulliam2000relationship}. The range of environmental variables (e.g. climate, land cover or soil characteristics) that matches the eco-physiological requirements of a species delimits its Grinnellian niche \citep{grinnell1917niche}. Habitat suitability models or species distribution models (SDMs) \citep{guisan2017habitat} aim to infer this niche by establishing statistical relationships between observed occurrences or abundances of species and the environmental (abiotic) characteristics of the corresponding locations. These models have been particularly useful to predict species in space and time \citep{ThuillerNatCom2019}, providing operational tools to conservation biologists \citep{guisan2013predicting,PollockTREE2021}. 

Beyond finding suitable habitats, living organisms meet their metabolic demands by feeding on, or acquiring, resources delimited by their Eltonian niche \citep{elton1927nature}. Through the processes of foraging for food or resources, reproducing and responding to the habitat conditions, species in a community affect each other, directly or via alterations of their surrounding environment (e.g. a large tree provides shade to shade-tolerant under-storey species). Moreover, species with shared resources may exclude one another locally (\textit{competitive exclusion} \citep{hardin1960competitive}) or be different enough in terms of space and resource needs to co-exist (\textit{niche partitioning} \citep{schoener1974resource}). Conversely, some species facilitate others by modifying the environment in a way that creates habitats or enables access to resources for other species (\textit{engineering and facilitation} \citep{cuddington2011ecosystem}). These biotic interactions can thus be symmetric (e.g. mutualism, competition) in some cases or asymmetric in many other cases (e.g. predator-prey interaction, amensalism, parasitism) \citep{morales2015inferring}. 
Although biotic interactions are deemed to take place locally, they are likely driving spatial variation in species abundances \citep{boulangeatELE2012}, and may alter species ranges and leave imprints at large spatial scales \citep{gotelli2002ecology, wisz2013role}, but see \citep{thuiller2015species}. 

As a result of species 'Grinnellian' and Eltonian's niches, together with species dispersal abilities, species co-abundances vary in space. These data, measured as \textit{community data}, are usually the corner-stone of analyses aiming to tease apart the relative importance of these processes \citep{weiher2001ecological,thuiller2013road,ovaskainen2017make}. A natural way to address this objective is to jointly model multiple species distributions against environmental variables, and then, analyse the pairwise co-dependencies between species after controlling for the environmental effects. In theory, these pairwise co-dependencies (i.e. associations) could represent \textit{the net effect} of one species on another, resulting from direct interactions or indirect effects \citep{ovaskainen2017make}. In practice, due to the intertwined effects of biotic and abiotic processes, they are also the outcome of model mis-specifications and errors,  of  missing environmental variables and interacting species \citep{PoggiatoTREE2021,blanchet2020co}. 

Several statistical frameworks have been proposed to infer these associations, either as their main objective or as a byproduct of the modeling process. These approaches differ in the type of dependencies they can model, in how they accommodate abundance data, and in the way they incorporate environmental covariates.  
Joint Species Distribution Models (JSDM, \cite{warton2015so}), the trendy tools at the moment, jointly predict the co-distributions of multiple species. Basically, once environmental covariates are accounted for, the residual correlation matrix is assumed to potentially capture species associations that are unexplained by the modeled covariates \cite{pollock2014understanding}. Recent implementations incorporate latent factors as a way to account for missing environmental variables and to reduce the parameter space size \citep{ovaskainen2017make}. Adaptation for abundance data, particularly counts, was achieved through either data transformation techniques or appropriate link functions \citep{clark2017generalized, niku2019gllvm, ovaskainen2020joint, chiquet2018variational,popovic2018general}. 
Alternatives mostly rely on Markov Random Fields (MRF) that can be applied to estimate conditional dependencies from a set of co-occurring species \citep{clark2018unravelling, harris2016inferring}, while accounting for the environmental variations. MRF have the statistical property of estimating direct associations between pairs of variables while accounting for all other associations, which makes them highly suitable in Ecology \citep{clark2018unravelling}. 

Although these two approaches and others have generated a renewed interest to understand biodiversity patterns from community data, they have also crystallized strong debates on their capacity at revealing true associations that can ultimately be linked to interactions. First, it has been shown that most implementations of JSDMs provide similar predictions and inferences than traditional SDMs since the residual correlation structure does not affect the estimated species-environment relationships \citep{PoggiatoTREE2021}. Second, simulated and empirical case studies have shown the difficulties of these approaches to infer simulated species associations \citep{zurellJSDM2018, konig}. Third, extracting species associations from co-occurrence data proves to be a complex, if not impossible, problem \citep{blanchet2020co, cazelles2016theory}. Last, but not the least, since both JSDM and MRF infer a precision matrix from the correlations between prediction residuals, they can only retrieve symmetric associations. This is not a desired properties since most species interactions, hence their induced associations, are likely to be asymmetric \citep{morales2015inferring}. 

Still, we believe that analysing community data along environmental gradients can bring useful information to infer species associations. To achieve such a long term goal, we need an approach that can handle both symmetric and asymmetric associations. Doing so requires capturing the way a given species affects the others, but also how the same species is affected by the other species. Interestingly, this duality has long been used in functional ecology to represent how a species respond to the environment through its 'response traits', and how it affects community functioning through its 'effect traits' \citep{lavorel2022}. An extension of this response-effect framework has been proposed for trophic interactions by linking response traits at a given trophic level to effect traits at another level \citep{lavorel2013trophic, gravel2016phyltrans}. We thus believe that distinguishing how a species respond to a species or a community from how it can affect it in return, which ultimately depend on the intrinsic properties or traits of the species, could provide a more suitable framework to make the best of community data and potentially extract information on species associations. 

\paragraph{Outline}
Here, we propose a framework that builds on this response-effect concept to model species - environment relationships and pairwise symmetric and asymmetric (i.e. bidirectional) associations all-together. To do so, we use machine learning tools to build an efficient dependency network \cite{heckerman2000dependency} encoding bidirectional species associations from community data. These associations are represented with two sets of embeddings encoding both species effects and responses to other species. Ultimately, the final conditional model of species abundances is  built by aggregating both species-environment relationships and the biotic embeddings through different implementations that can best represent mechanistic understanding of the system (e.g. predator-prey interactions, competition-facilitation-amensalism-comensalism). 

Through two experiments on simulated datasets and an empirical case study, we illustrate the different implementation of the interplay of biotic associations with environmental covariates. First, we simulate species abundance data with a species community model to evaluate the ability of our framework to recover known associations (both symmetric and asymmetric), where we assume an additive partitioning of environmental and biotic filters, and compare it with state-of-the-art joint species distribution models and Markov-random fields. Second, we evaluate the ability of our model to recover simulated predator-prey associations under different food web topologies, assuming a multiplicative effects of environmental and biotic filters. 

Finally, we apply our framework to a well-studied Alpine plant community dataset \cite{choler2005consistent,warton2015so} representing an example of hierarchical filtering of assembly rules (environmental effects at regional scale and competitive-facilitate interactions at the community scale). We used this empirical example to illustrate the analysis of structure of the species association networks.

\newpage
\section{The framework}

\begin{figure}[htbp]
	\begin{subfigure}{0.45\textwidth}
		\includegraphics[scale=.8]{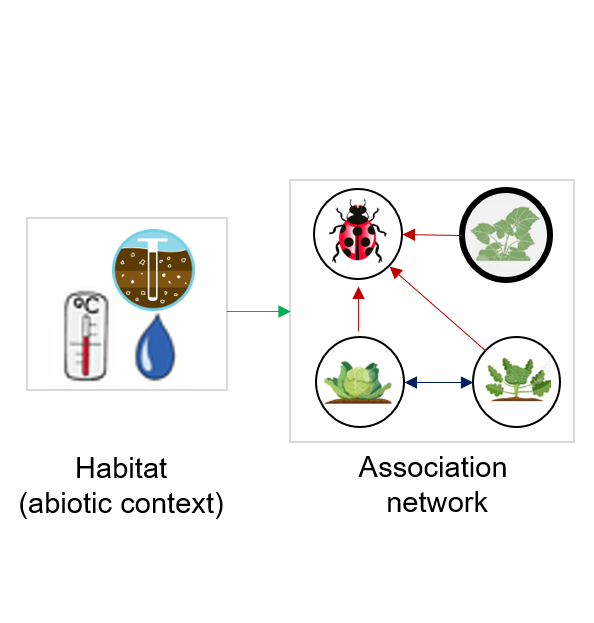} 
		\caption{Overall graphical model.}
		\label{fwk:a}
	\end{subfigure}
	\hspace{1em}
	\begin{subfigure}{0.35\textwidth}
		\includegraphics[scale=.8]{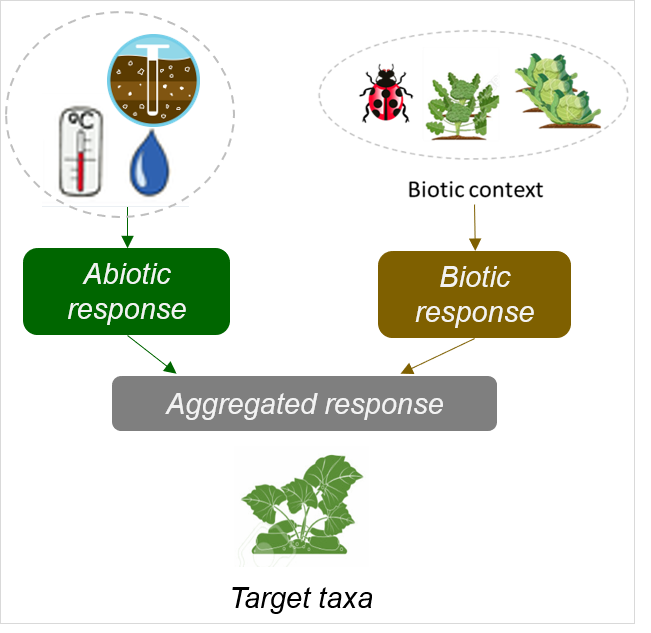} 
		\caption{Node-wise neighborhood inference.}
		\label{fwk:b}
	\end{subfigure}
	
	\caption{A graphical illustration of the interplay between the environmental and the biotic filters. (a) Species in a community form a network of associations of different signs: positive (red), negative (blue). Each edge represents an association from a source to a target species. When both species influence each other, the association is bidirectional. All species, along with their associations, respond to the environmental conditions (green). We ignore the reciprocal effect of species on the environment. (b) The abundance of a given species results of its aggregated response to the environment and to the biotic contexts. } \label{fwk}
\end{figure}

At a high level, our framework models both species–environment responses and species–species associations. It captures how species respond to environmental conditions and to other species (i.e., species response), as well as how they influence the abundance or occurrence of others (i.e., species effect). The resulting graphical model is directional with respect to the environment (which drives species abundance, but not vice versa), and bidirectional with respect to species associations (~Fig\ref{fwk:a}).

More specifically, the incoming edge weights for each species are estimated through a multiple regression that includes both environmental covariates and the abundances of co-occurring species (~Fig\ref{fwk:b}). By incorporating all potential predictors in a single model, the framework quantifies \textit{conditional dependencies} and disentangles environmental effects from biotic ones, allowing a direct assessment of their relative contributions. To represent bidirectional associations, we learn separate embeddings for how a species influences others and how it is influenced by them.

In the following, we define these embeddings and describe the conditional abundance model and its implementation.

\paragraph{Notation}
We consider a site by species matrix ($\boldsymbol{\siteS}$, $\boldsymbol{\spcS}$), that contains the abundance of species $\spc$ at site $\site$, denoted $\abdv_{\site\spc}$. At each site $\site$, the vector $\boldsymbol{\abiov_{\site}}$ represents the environmental covariates.

\subsection{Spatial associations and biotic context}
\subsubsection{Representing species associations using embeddings}
\begin{figure}[htbp]
	\begin{subfigure}{0.4\textwidth}
		\includegraphics[page=2,scale=1]{tikz_figures.pdf} 
		\caption{Association matrix factorization.}\label{assocdomain:a}
	\end{subfigure}
	\hspace{1em}
	\begin{subfigure}{0.4\textwidth}
	    \includegraphics[page=1]{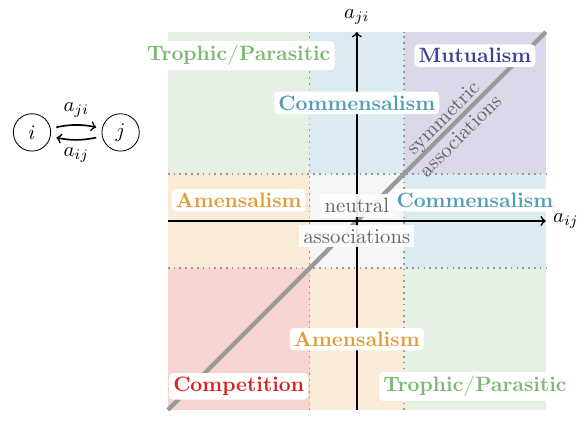}
	\caption{Pairwise associations classification}\label{assocdomain:b}
	\end{subfigure}
	\caption{Association strengths are computed from species response and effects (a). Pairwise association strengths are mapped to potential interaction classes (b). The different quarters of the bi-plot represent the various types of associations between species. The 1:1 line represents symmetric associations.} \label{assocdomain}
\end{figure}

For a given pair of species, a \textit{spatial association} describes the statistical influence of a species on the abundance of another species. The influences can be of different polarity (positive, negative or neutral) and have different intensities (Fig~\ref{assocdomain:b}). Several mechanisms can lead to these associations: a direct interaction (e.g.\ pollination, predation), an indirect interaction through the environment (e.g.\ resource competition) or a shared correlation to an unmeasured environmental variable or unobserved species \citep{PoggiatoTREE2021}.

We note $\Assv_{\spc\ospc}$ the influence of species $\ospc$ on species $\spc$ which represents the change in abundance (excess if
positive, deficit if negative) of a \textit{target}
species $\spc$ induced by a \textit{source} species $\ospc$.  These values that represent the sign and the strength of the association between all species pairs are stored into an $m \times m$ asymmetric association matrix $\boldsymbol{\AssMat}$ (\ Fig~\ref{assocdomain:a}).

From a response-effect perspective, any element of $\boldsymbol{\AssMat}$ (e.g. $\Assv_{\spc\ospc}$) is the byproduct of the effect of species $\ospc$ and the response of species $\spc$. We assume that these parameters represent latent traits or properties of the species that we do not observe, and which we implement  as two separate $d$-dimensional embeddings. 

The \emph{effect embedding} of species $\spc$, $\boldsymbol{\Effv_\spc}$, expresses the type (i.e. traits, properties) of organisms the species allows or impedes when it is present. The \emph{response embedding} of species $\spc$, $\boldsymbol{\Resv_\spc}$, expresses the type (i.e. traits, properties) of biotic context it can tolerate. For instance, trees with dense canopy create shade (effect) that selects only shade-tolerant (response) species and exclude others.

The response and effect embeddings of the different species are collected into two $m \times d$ matrices, respectively denoted as $\boldsymbol{\ResMat}$ and $\boldsymbol{\EffMat}$. The association matrix is then written as %
$\boldsymbol{\AssMat} = \boldsymbol{\ResMat\EffMat}^{T}$ (cf.\ Fig~\ref{assocdomain:a}).

\subsubsection{Biotic context}
The biotic context encodes our assumptions about the potential effects a target species is exposed to at a given site. In the simplest case, without any prior knowledge, it consists of the abundance of other species observed at the same site. Formally, the biotic context of species $\spc$ at site $\site$, denoted $\BioCont_{\site\spc}$, is defined as follows:

\begin{equation*}
\BioCont_{\site\spc} = \{\ospc \in \spcS, \ospc \neq \spc ~\text{and}~ \abdv_{\site\ospc} > 0 \}\,.
\end{equation*}

We obtain the aggregated effect of the biotic context by averaging the effect embeddings of its elements weighted by species' abundances: 

\begin{equation*}
\BioEff_{\site\spc} = \frac{1}{\abs{\BioCont_{\site\spc}}} \sum_{\ospc \in \BioCont_{\site\spc}} \abdv_{\site\ospc} \Effv_{\ospc}\,.
\end{equation*}

This formulation allows the presence of opposing effects from different species to balance one another. 

The biotic context constrains the structure of the inferred species association network by restricting the set of potential associations \textit{a priori}. For instance, it can be easily adapted for each species according to known interactions. It can also include species from neighboring locations (\textbf{spatially-explicit}) up to a chosen radius within which their influence would be considered relevant (e.g.\ species with low/high mobility). Similarly, we can construct a \textbf{temporally-explicit} biotic context from previous observations to account for time-lag and phenological mismatches. 
(See Supplementary Methods for other biotic context formulation variants).

\subsection{A conditional generative model of abundance}

\subsubsection{Conditional generative model}
To disentangle environmental and biotic effects, we represent the response of the species $\spc$ at site $\site$ as an aggregation function $f_{agg}$ of the environmental $\eta_{\site\spc}^A$ and biotic $\eta_{\site\spc}^B$ responses (see Eq.~\eqref{eq:1a}). The environmental response is given by the habitat suitability model(s) $\habs_{\spc}$ involving only the environmental covariates $\abiov_{\site}$ (see Eq.~\eqref{eq:1b}). The biotic response $\eta_{\site\spc}^B$ depends on the response embedding $\Resv_\spc$ of the target species and on the biotic context effect $\BioEff_{\site\spc}$ resulting in an abundance-weighted sum of pairwise association strengths (see Eq.~\eqref{eq:1c}). An offset $o_{\spc}$ is used to account for variation in exposure or effort. 

The response of each species conditional to the environmental conditions and biotic context ($\abdv_{\site\spc} \mid {\abiov_\site,\BioCont_{\site\spc}})$, denoted as $\abdv_{\site\spc}$ for clarity, is assumed to follow a distribution $\mathcal{F}$ from the Exponential Family with mean $m_{\site\spc}$  and dispersion $\phi_\spc$ parameters. The function $g$ denotes the canonical link function, which relates the aggregated response at site $\site$ of species $\spc$ to the mean (see Eq.~\eqref{eq:1d}).

The choice of distribution within this family is done according to the data type, for instance, using the normal distribution for biomass, the Bernoulli for presence/absence, or the Negative Binomial for over-dispersed counts.

\begin{subequations}\label{eq:1}
	\begin{align} 
	g(m_{ki}) = f_{agg}(\eta_{\site\spc}^A,\eta_{\site\spc}^B) \label{eq:1a}\\
	\eta_{\site\spc}^A = \habs_{\spc}(\abiov_{\site}) \label{eq:1b}\\
	\eta_{\site\spc}^B = o_\spc + \Resv_\spc \BioEff_{\site\spc} =  o_\spc + \sum_{\ospc \in \BioCont_{\site\spc}} (\abdv_{\site\ospc} * \Assv_{\spc\ospc}) \label{eq:1c}\\
    \abdv_{\site\spc} \sim \mathcal{F}(m_{ki}, \phi_i)\label{eq:1d} 
	\end{align}
\end{subequations}

\subsubsection{Aggregation of abiotic and biotic effects}
\begin{figure}[bthp]
		\includegraphics[scale=.5]{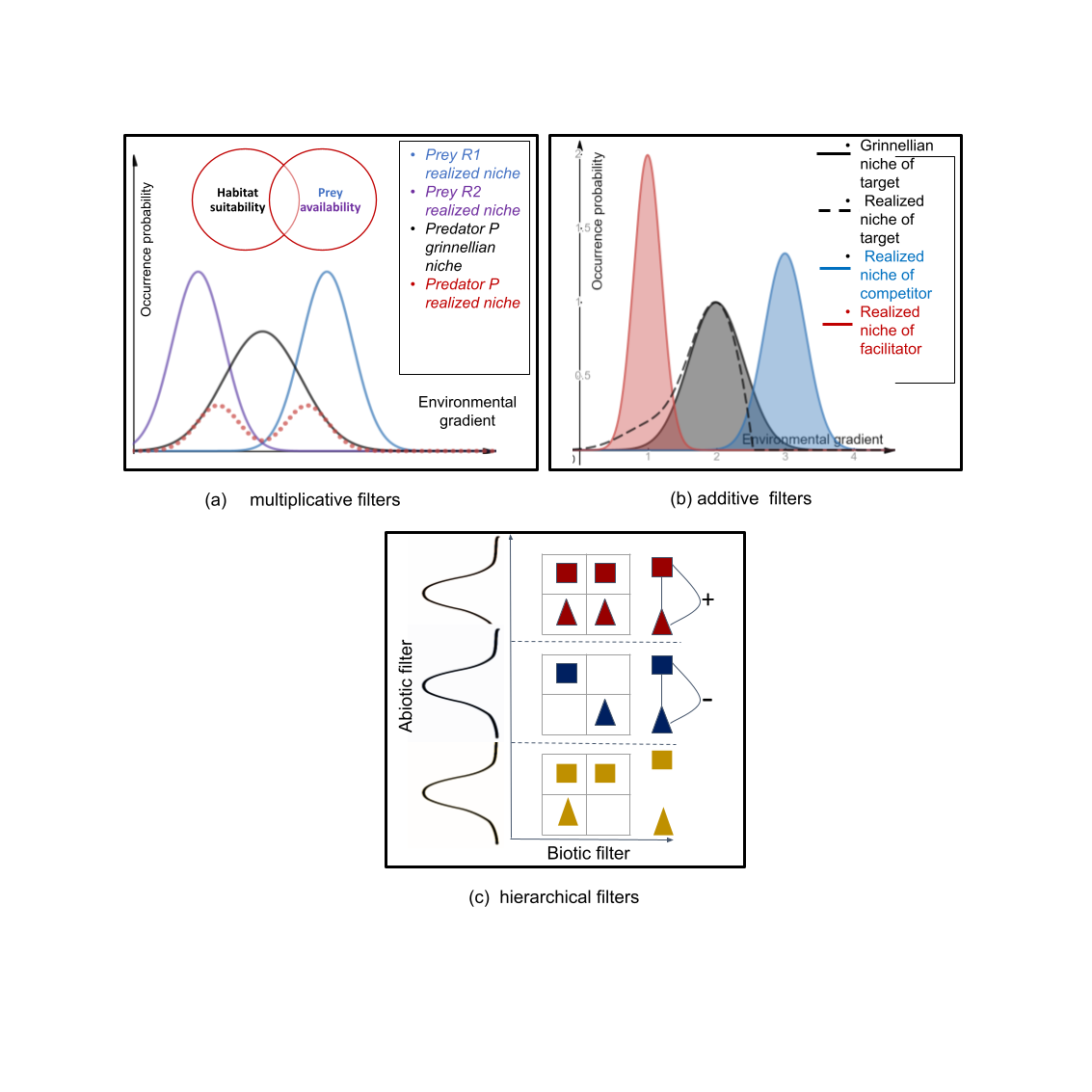} 	
	\caption{Examples of scenarios for the aggregation of abiotic and biotic filters. (a) multiplicative filters represent a product of: environmental requirements and biotic resources (realized niche of preys). (b) Additive filters represent the sum of three complementary factors: the effect of the environmental environment and the associations that complement it through either presence of facilitators (resp. competitors) that can extend (resp. restrict) the suitable environmental range. (c) Hierarchical filters show the effect of two nested filters. First, an environmental filter operates at a regional scale, as depicted in the vertical stratification of species into 3 groups (regional pools): red, blue and yellow. Second, a biotic filter that acts locally through the presence of positive or negative associations or lack thereof between species of the regional pool.} \label{config}
\end{figure}

The aggregation function captures the interplay between abiotic and biotic filters (Fig.~\ref{config}), which can follow different ecological assumptions. In the community assembly rule framework \cite{weiher2001ecological}, environmental conditions first define the set of species that can potentially occur, while biotic interactions determine which of these species persist, based on their responses and effects. In some cases, abiotic conditions condition the occurrence of an interaction or share its nature or strength. In others, the biotic context itself can create favorable conditions, for instance, through facilitation.

\noindent To reflect these possibilities, we implement and evaluate three aggregation modes: additive, multiplicative, and hierarchical. Although these capture distinct ecological mechanisms, the framework is flexible and can accommodate alternative formulations (Fig.~\ref{config}).

\paragraph{Additive filters}
\begin{align}
   g(m_{\site\spc}) = \eta_{\site\spc}^A + \eta_{\site\spc}^B \label{agg:add}
\end{align}

In the case of additive filters, the biotic context can complement the environmental conditions, and a species may occur if either filter is favorable. For instance, a species might be present even in unsuitable habitat if another facilitator species creates favorable micro-habitat. Conversely, a competitor’s presence might exclude a species despite suitable environmental conditions. Here, the biotic and environmental responses are summed to reflect their combined, potentially compensatory influence (Eq.\ref{agg:add}).

\paragraph{Multiplicative filters}
\begin{align}
    m_{\site\spc} = \sigma(\eta_{\site\spc}^A) \times \sigma(\eta_{\site\spc}^B)\label{agg:mul} \\
    \abdv_{\site\spc} \sim Bernoulli(m_{\site\spc})\label{agg:muly}
\end{align}

In the case of multiplicative effects, a species can only be present when both abiotic conditions and the biotic context are favorable. This setting is particularly relevant for obligate interactions, such as trophic, host–parasite, or host–symbiont relationships. For example, consider a predator species that requires both suitable abiotic conditions (e.g., temperature) and the presence of at least one prey species. Its occurrence (Eq \ref{agg:muly}) depends on the product of these two components (Eq \ref{agg:mul}), each assessed independently, reflecting the necessity of both filters for survival.

\paragraph{Hierarchical filters}

The hierarchical filter setting distinguishes between two nested levels of influence: broad-scale environmental filters (e.g., climate) that define the regional species pool, and local-scale habitat features that interact with biotic associations to shape species abundances. This approach mirrors the commonly used assembly rule framework \citep{thuiller2013road}, in which abiotic filters act first, followed by biotic structuring. In practice, we implement this structure using a zero-inflated regression, where the environmental component governs species presence (denoted by the suitability $s_{\site\spc}$) (Eq \ref{agg:hiersdm}), and the biotic context influences the abundance $\abdv_{\site\spc}$ conditional on occurrence. In this formulation, a Dirac point mass at zero is used such that when $s_{\site\spc} = 0$, the species abundance is deterministically set to zero (Eq \ref{agg:hierab}).

\begin{align}
  s_{\site\spc} \sim Bernoulli(\eta_{\site\spc}^A)\label{agg:hiersdm}\\
  \abdv_{\site\spc} \sim \begin{cases}
    \mathcal{F}(\eta_{\site\spc}^B ; \phi_\spc), & \text{if $s_{\site\spc}=0$}.\\
    \delta_0, & \text{otherwise} \label{agg:hierab}.
  \end{cases}
\end{align}

\hfill \break
Choosing the aggregation function requires knowledge of the ecological community and the expected types of interactions or dependencies that could induce the inferred associations. In contrast, different assumptions can be tested and the best one can be quantified by statistical model selection.


\subsection{Inference and model selection}
We use the Stochastic Gradient Descent algorithm \citep{bottou2010large} to optimize the negative log-likelihood of the observed abundances or occurrences with respect to the parameters, including the response and effect embedding matrices, the parameters of the abiotic response weights, and the species-specific dispersion parameters. Since the biotic context can substantially increase the number of variables and thus the risk of variance inflation due to multicollinearity \citep{dormann2013collinearity}, we introduce elastic net regularization penalties to select meaningful associations for each species.
 
The proposed model includes a set of hyperparameters that must be selected carefully: the hyperparameters for the habitat suitability models, the embedding dimension, the vector of species offsets, and the regularization coefficient. We implemented two model selection (hyperparameter tuning) strategies. The first relies on information criteria \citep{konishi2008information} to penalize model complexity, such as the Akaike Information Criterion (AIC), the Bayesian Information Criterion (BIC), or its extended version (eBIC). The second strategy uses cross-validation based on predictive performance, using, for instance, the AUC for presence/absence data or Poisson deviance for count data. 

\subsection{Species association network}
The regularization introduces sparsity into the association network by removing links between species that are independent or do not exhibit strong associations \citep{ohlmann2018mapping}. Alternatively, the robustness of estimated biotic associations can be assessed through a bootstrap procedure: instead of applying a penalty during inference, the model is fitted to multiple bootstrap samples of the original dataset. Confidence intervals for the mean of each pairwise association are then computed, and associations whose intervals include zero are set to zero. In both the regularized and bootstrap approaches, a threshold can be applied to obtain a discrete version of the association matrix, defined as follows:

\begin{equation*}
\InterMat_{\spc\ospc} =  \left\{
\begin{array}{ll}
\text{positive} & \mbox{if } \Assv_{\spc\ospc}>\epsPos, \\
\text{negative} & \mbox{if } \Assv_{\spc\ospc}<-\epsNeg, \\
\text{neutral} & \mbox{otherwise.}
\end{array}
\right.
\end{equation*}

such that $\epsPos$ and $\epsNeg$ represent user-defined thresholds on the strength of the positive and negative associations, respectively. The resulting matrix can be seen as a network, where each species is represented by a vertex and a directed edge labeled as positive (resp.\ negative) from vertex $\spc$ to vertex $\ospc$ represents a positive (resp.\ negative) influence of species $\spc$ on species $\ospc$.

Based on our embedding definitions, species with similar response embeddings form clusters of rows in the association matrix, referred to as \textit{response groups}, while species with similar effect embeddings form clusters of columns, or \textit{effect groups}. These two sets of groups can be identified simultaneously using a co-clustering algorithm \citep{govaert2013co}. Their combination reveals blocks in the association matrix corresponding to groups of species that play similar structural roles i.e., are functionally redundant or exchangeable within the network \citep{gauzens2015trophic}.
\section{Test of the framework on simulated species communities}

To validate our framework, we conducted two simulation experiments in which community data were generated along an environmental gradient based on species-specific abiotic optima and predefined association matrices.

The first experiment was designed to assess the ability of our model (EA) and competing association inference methods (JSDMs, MRFs) to recover both symmetric and asymmetric associations under an additive filtering scenario.

The second experiment was designed to test whether our model could recover associations under a multiplicative filtering scenario, where species presence depends on both environmental and biotic context suitability. This setting is not supported by the alternative methods.

\subsection{Experiment 1: process-based simulation of community assembly}
\subsubsection{Community data simulation}
We used a process-based stochastic model adapted from \texttt{Virtualcom} \cite{munkemuller2015virtualcom} to simulate the assembly of individuals from a regional species pool into communities, on different locations sampled along an environmental gradient (See Supplementary Methods).

\noindent We designed an experiment in which multiple simulations were run on random locations along a single environmental gradient (ranging from 0 to 100), using randomly generated configurations of the prior association matrix. These configurations included:
\begin{itemize}\setlength\itemsep{0em} \setlength\parsep{0em}
    \item Only environmental filtering (Env)
    \item Only positive associations (Pos)
    \item Only negative associations (Neg)
    \item Mixed positive and negative associations (PosNeg)
\end{itemize}

For each configuration, we varied:
\begin{itemize}\setlength\itemsep{0em} \setlength\parsep{0em}
    \item The species pool size: 10, 20, or 50 species
    \item The association density: sparse ($1/3$ of species pairs associated) vs. dense ($2/3$)
    \item The association symmetry: symmetric (+/+, -/-) vs. asymmetric associations (+/0,-/0).
\end{itemize}

\noindent Association strengths were fixed at $+1$ for positive and $-1$ for negative effects, focusing on association polarity rather than intensity. This factorial design yielded 33 simulation datasets, allowing us to evaluate our framework across a range of conditions and compare its ability to recover symmetric associations against JSDMs and probabilistic graphical models.

\subsubsection{Inference}
We fitted our model to species count data from the simulated communities using a negative binomial distribution with an exponential link function and an additive aggregation of environmental and biotic filters. Hyperparameters were selected via 10-fold cross-validation, using Poisson deviance as the performance metric.

\subsubsection{Evaluation and comparison with JSDMs and graphical models}
We also applied five well-established or emerging methods for inferring associations from count data: HMSC \citep{ovaskainen2017make}, EcoCopula \citep{popovic2019untangling}, EMTree \citep{momal2019tree}, MRFcov \citep{clark2018unravelling}, and PLN \citep{chiquet2018variational}. Table~\ref{tab:models} summarizes, for each method, the underlying probabilistic model, data requirements, training/inference settings, and any additional post-processing steps.

For all methods, the inferred association matrices were discretized to identify the type of association (positive, negative, or neutral), and compared to the ground-truth using standard multi-class performance metrics: precision, recall, and F1-score. Recall reflects the proportion of true associations of a given type that were correctly identified (sensitivity), while precision measures the proportion of predicted associations of a given type that were correct (specificity). The F1-score is the harmonic mean of precision and recall, balancing false positives and false negatives.

\begin{table}[hbtp]
\resizebox{\textwidth}{!}{%
\begin{tabular}{|l|l|l|l|l|}
\hline
\rowcolor[HTML]{C0C0C0} 
\multicolumn{1}{|c|}{\cellcolor[HTML]{C0C0C0}\textbf{Framework}} &
  \textbf{Count distribution} &
  \multicolumn{1}{c|}{\cellcolor[HTML]{C0C0C0}\textbf{Association structure}} &
  \textbf{Graph selection procedure} &
  \multicolumn{1}{c|}{\cellcolor[HTML]{C0C0C0}\textbf{Learning configuration}} \\ \hline
\begin{tabular}[c]{@{}l@{}}Ecological \\ Association \\ Network (EA)\end{tabular} &
  \begin{tabular}[c]{@{}l@{}}-Negative binomial\\ -Unknown dispersion\\ -Link: log\end{tabular} &
  \begin{tabular}[c]{@{}l@{}}Dependency \\ network\end{tabular} &
  Cross-validation &
  \begin{tabular}[c]{@{}l@{}}Optimizer: adam\\ Maximum number of epochs: 200 \\ Batch size: 16\\ Early stopping (convergence \\ by monitoring validation loss):\\ - Patience: 5 epochs\\ - Tolerance: 1E-3\end{tabular} \\ \hline
EcoCopula \cite{popovic2018general,popovic2019untangling} &
  \begin{tabular}[c]{@{}l@{}}-Negative binomial\\ -Unknown dispersion\\ -Link: log\end{tabular} &
  \begin{tabular}[c]{@{}l@{}}Copula Gaussian \\ Graphical Model\end{tabular} &
  Graphical lasso &
  Importance sampling: 1000 \\ \hline
EMtree \cite{momal2019tree} &
  Poisson log-normal &
  \begin{tabular}[c]{@{}l@{}}Mixture of tree-shaped Gaussian \\ Graphical Models\end{tabular} &
  \begin{tabular}[c]{@{}l@{}}Edge probabilty\\ Support: 2/pool\_size\end{tabular} &
  \begin{tabular}[c]{@{}l@{}}Covariance mode: full\\ Number of iterations: 50\\ \\ Convergence tolerance: 1E-8\\ Resampling: 5\end{tabular} \\ \hline
\begin{tabular}[c]{@{}l@{}}Hierarchical Modeling \\ of  Ecological Communities\\ (HMSC) \cite{ovaskainen2017make}\end{tabular} &
  Poisson &
  Residual correlation &
  \begin{tabular}[c]{@{}l@{}}MAP residual covariance\\ Support level 95\%\end{tabular} &
  \begin{tabular}[c]{@{}l@{}}MCMC: Hamiltonian Monte-Carlo\\ thinning = 10\\ nChains = 2\\ Burn-in = 500\\ nSamples = 5000-7500\\ \\ Cross-validation: 2 folds\end{tabular} \\ \hline
MRFcov \cite{clark2018unravelling} &
  \begin{tabular}[c]{@{}l@{}}Gaussian with \\ non-paranormal \\ transformation \\ of count data\end{tabular} &
  \begin{tabular}[c]{@{}l@{}}Conditional Markov \\ Random Field\end{tabular} &
  \begin{tabular}[c]{@{}l@{}}Bootstrap (95\% CI)\\ Sample proportion: 70\%\\ Symetrization function: mean\end{tabular} &
  Bootstrap samples: 500 \\ \hline
\begin{tabular}[c]{@{}l@{}}Poisson\\ Log-Normal network\\ (PLNetwork) \cite{chiquet2018variational}\end{tabular} &
  Poisson log-normal &
  \begin{tabular}[c]{@{}l@{}}Gaussian Graphical \\ Model\end{tabular} &
  \begin{tabular}[c]{@{}l@{}}Graphical lasso + \\ stability selection\end{tabular} &
  \begin{tabular}[c]{@{}l@{}}Covariance mode: full\\ Offset: None\end{tabular} \\ \hline
\end{tabular}%
}
\caption{Description of the evaluated frameworks and their respective configuration.}
\label{tab:models}
\end{table}

\subsubsection{Results}

The analysis of the relative abundance index (\evRAI{}) showed good discrimination between positive and negative associations, while neutral associations resulted in more variable \evRAI{} values (see Supplementary Materials).

\noindent Among the six inference methods, some were relatively easy to fit and offered limited control over model selection (e.g., EMTree, EcoCopula, MRFcov), beyond setting the number of iterations and the sampling scheme used for estimating confidence intervals. Execution time varied considerably between methods, largely driven by the model selection procedures employed (e.g., bootstrapping, lasso regularization, cross-validation). In particular, the Bayesian posterior inference in HMSC made it substantially slower than the other approaches.

\begin{figure}[bthp]
	\centering
	\includegraphics[scale=.75]{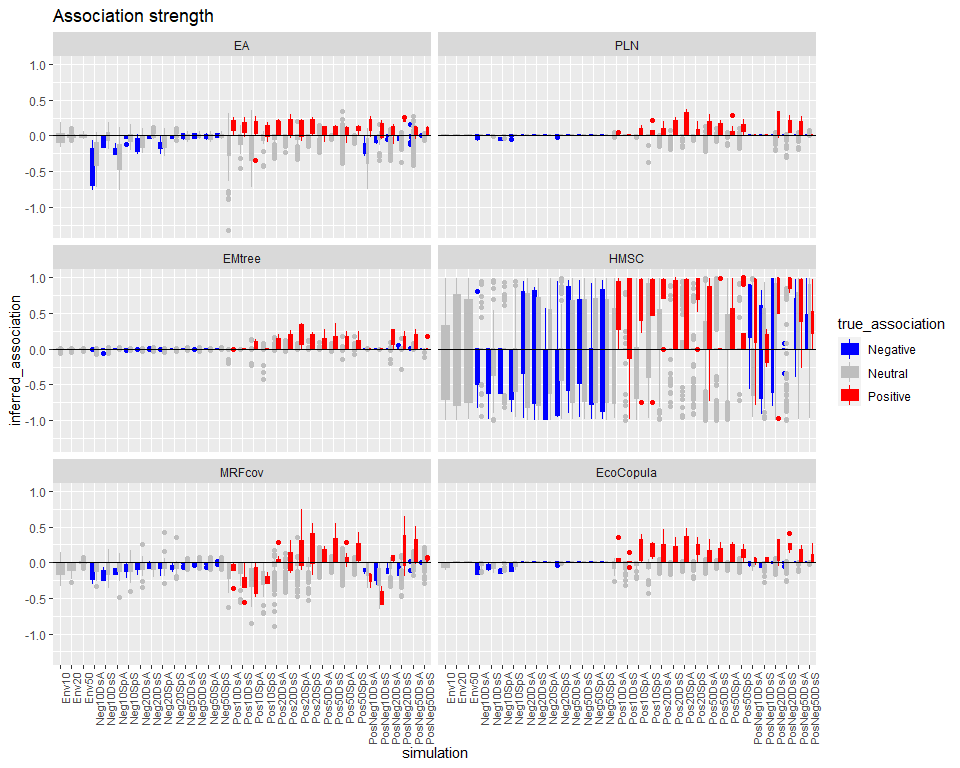} 	
	\caption{Distribution of the association strengths inferred by the six methods run for each simulation experiment (in columns, e.g. Neg\_S\_10\_D means simulated \textbf{Neg}ative \textbf{S}ymmetric associations for a species pool of \textbf{10} species, with \textbf{D}ense network of interactions). A data point represents a directed association from a species to another, its color encodes the true type of the association, its coordinates on the y axis represents the fitted association strength by the corresponding model for the simulation configuration x axis. In other words, blue histograms should rather be on the negative side (negative association correctly inferred), red colors on the positive side (positive association correctly inferred) and grey centered around 0 (neutral association)} \label{exp1strength}
\end{figure}

\noindent For each method, we visualized the distribution of inferred association strengths across the different simulation configurations (e.g., environmental filtering only, association types, species pool size; Fig.~\ref{exp1strength}). 

\noindent All methods except HMSC produced sparse association networks, with low strengths values and were good at discriminating positive and negative associations, while maintaining neutral associations median-centered at zero. 

\noindent Most spurious associations, i.e.\ neutral pairs with inferred value significantly different from zero, were negative especially in simulations involving only positive associations reflecting the implicit constraint induced by the fixed carrying capacity on the total species count. On the other hand, HMSC produced very dense association matrices despite a large support level for association selection suggesting that some of the inferred associations are indirect associations. There was no difference in inferred strengths neither between symmetric and asymmetric simulations (for all methods).

\noindent As species pool size increases and niche overlap becomes more likely, inferred association strengths become more sensitive to niche differences: positive associations weaken with increasing overlap across all methods, while negative associations show variable responses, with EA and HMSC displaying opposite trends as niche distance increases. 

\noindent In terms of association type classification, no major performance differences were observed across models for symmetric or dense association structures, the quality of inferred associations depended on species pool size: EA and EcoCopula consistently performed best on positive associations, especially in small pools, while negative associations were challenging for all methods, with EA, MRFcov, and HMSC showing relatively better performance. 

\noindent Detailed results are presented in the Supplementary Materials.

\newpage
\subsection{Experiment 2: simulation of predator-prey co-occurrences} 

\subsubsection{Community data simulation}

In this experiment, we simulated species occurrence data where each species depends jointly on suitable environmental conditions and the presence of at least one prey given by a known predator-prey network (food web). This setup reflects an intersection of abiotic and biotic filters, modeled as a multiplicative response. 
 
Using the \texttt{trophic} R package, we generated six food webs with different topologies, each involving the same number of trophic groups ($G = 5$). A trophic group consists of species that share the same prey and are consumed by the same predators, statistically analogous to a latent block structure in a graph. 

To each trophic group, we assign $m_G=5$ species with different abiotic niche optima sampled uniformly along an environmental gradient ranging from 0 to 100. We select 500 sites uniformly in the same gradient.  

\subsubsection{Inference}
We fitted our model to the simulated presence/absence data using a multiplicative filter setting. We used a linear logistic regression with a quadratic term to fit the Gaussian abiotic niche. The simulation model assumes a unidirectional positive dependency of the predators on their preys. Thus, we imposed a non-negative constraint to the embedding vectors to prevent the inference of negative associations and promote sparsity of the association matrix \cite{hoyer2004non}. Consequently, we only inferred two types of associations: positive and neutral. Additionally, we tested whether imposing structure by sharing embeddings between species of the same trophic group improved the ability of the model to retrieve true potential and realized associations. 

We used a 10-fold cross-validation to select the combination of embedding dimension and lasso regularization that maximized the accuracy of predicted occurrences.  

\subsubsection{Evaluation}
We evaluated the quality of the recovered associations in terms of accuracy, ROC-AUC, sensitivity, and specificity. As ground truth, we used two reference food webs: (1) the potential food web (metaweb), which includes all possible interactions; and (2) the realized food web, obtained by filtering the metaweb to retain only interactions between species that co-occur at least in one site.

\subsubsection{Results}
\begin{figure}[bthp]
	\centering
	\includegraphics[scale=0.5]{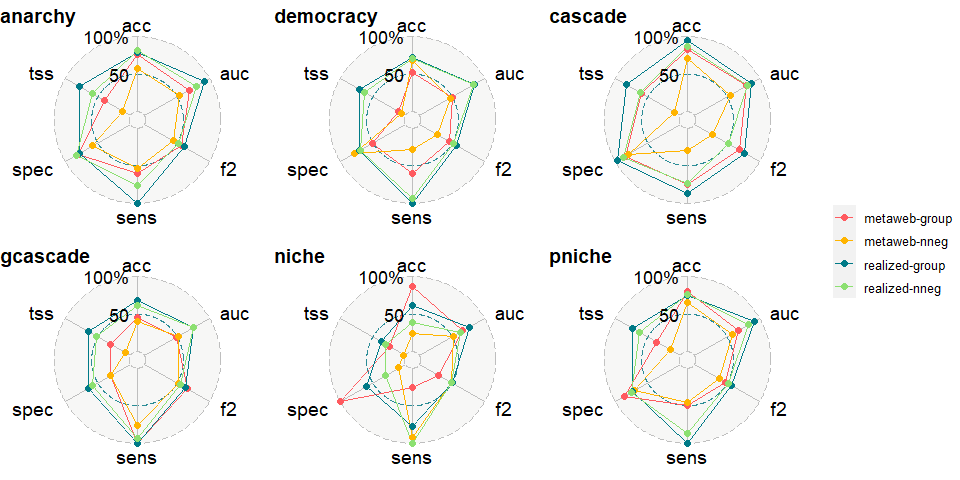} 	
	\caption{Network structure inference quality with respect to the potential (metaweb) and the realized networks under two different constraints: non-negative associations and within-group embedding sharing. Performances are reported separately for each food web topology.} \label{fwperf}
\end{figure}

The inferred associations were more faithful to the realized than the potential network (Fig ~\ref{fwperf}). In all cases, incorporating a parameter-sharing constraint within trophic groups allowed to improve the sensitivity with respect to both ground truth networks. 

Additional results are available in the Supplementary Materials.

\section{Empirical case study - Alpine plant associations}
\subsection{Plant community data}

To test our model on a real ecological system, we applied it to an Alpine plant abundance dataset originally published by \citet{choler2005consistent}. The dataset includes abundance records for 82 plant species surveyed in July 2000 across 75 vegetation plots (each $5 \times 5$ m) along a meso-topographical gradient in the French Alps. Environmental and topographic variables were also recorded for each plot.

\subsection{Inference and statistical analyses}
We fitted our model to this dataset using the hierarchical filtering mode (Fig.~\ref{config}), assuming that habitat suitability drives species occurrence, while local biotic associations influence species abundance and can lead to local exclusion \citep{boulangeat2012accounting}. Details of data pre-processing and model selection are provided in the Supplementary Materials.

\subsection{Results}
\begin{figure}[H]
	\begin{subfigure}{\textwidth}
		\centering
		\includegraphics[scale=0.15]{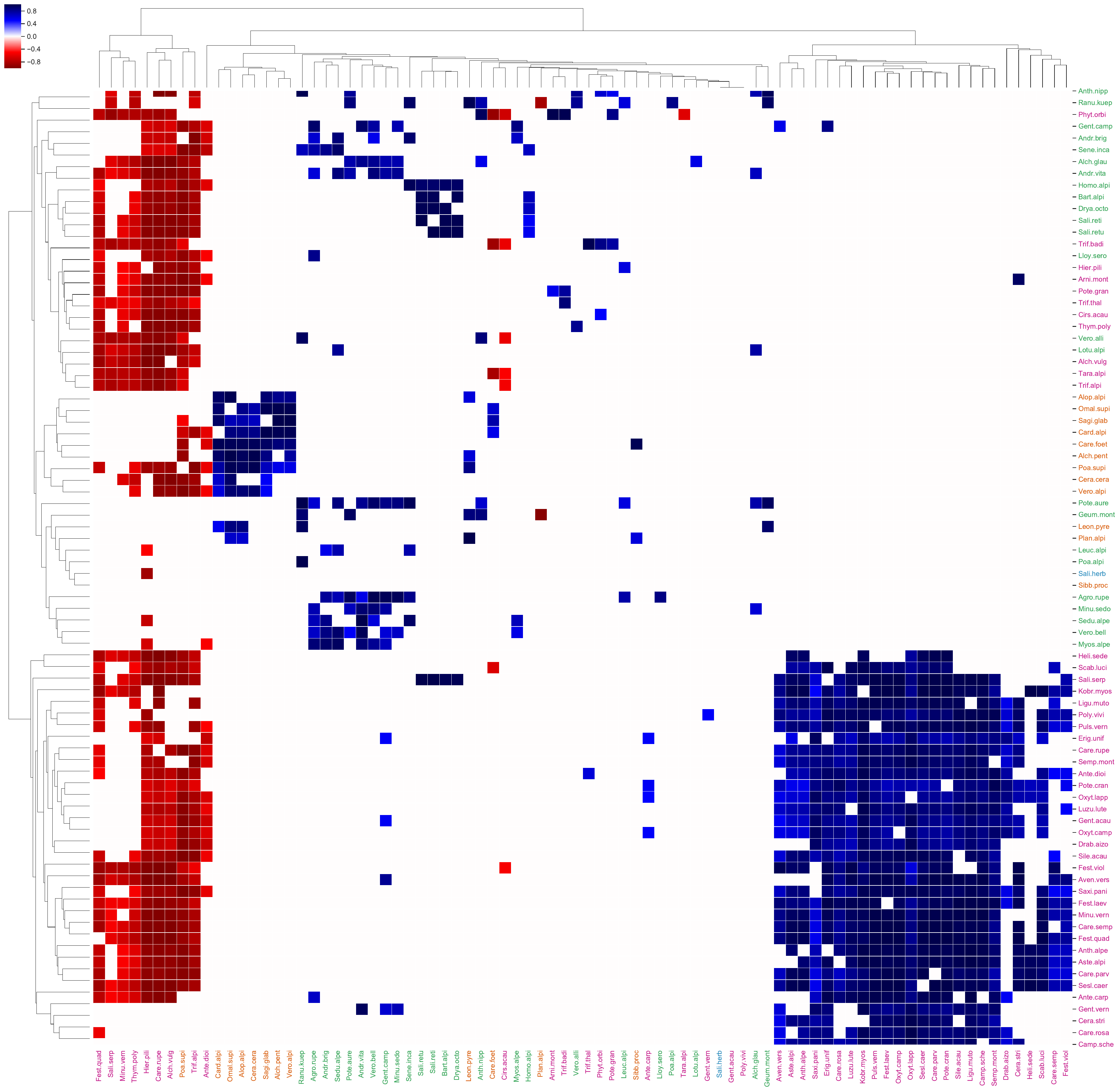}
		\caption{Inferred plant association matrix. Species in the association matrix are grouped based on a hierarchical co-clustering performed row-wise (yielding response groups) and column-wise (yielding effect groups).} \label{assocplant:a}
	\end{subfigure}
	\newline
	\centering
	\begin{subfigure}{\textwidth}
		\includegraphics[scale=0.4]{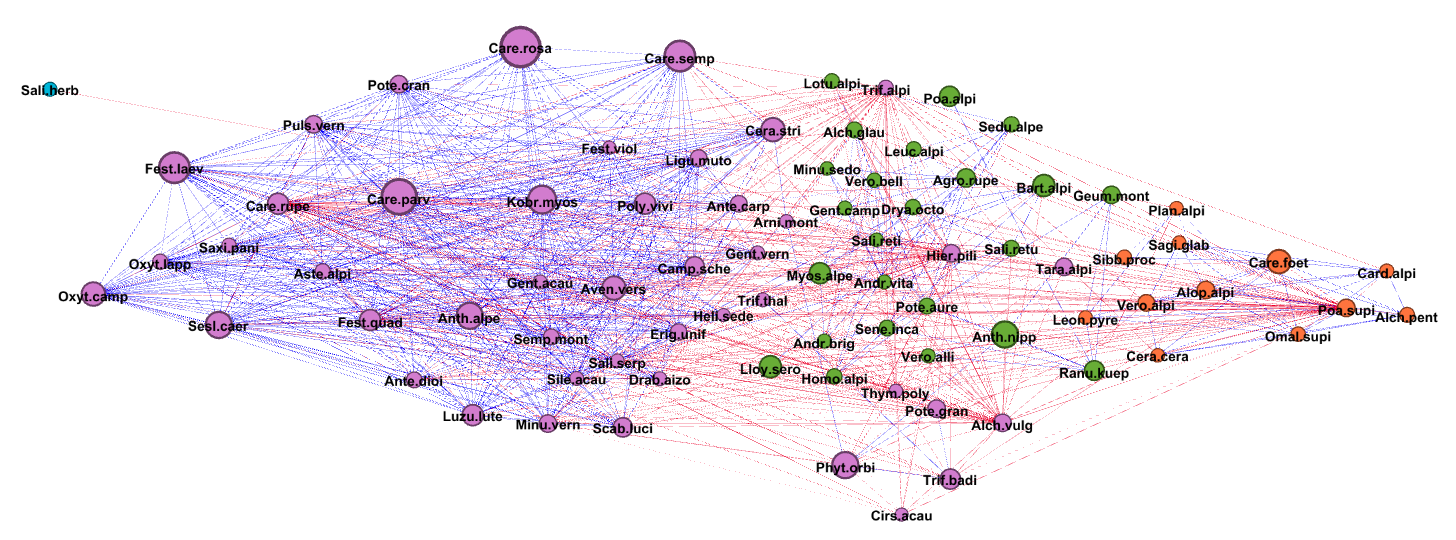}
		\caption{Network of plant associations. Blue (resp.\ red) edges indicate negative (resp.\ positive) edge weights. Node colors on the graph represent communities identified by the modularity maximization algorithm \cite{newman2006modularity} whilst node sizes are scaled according to the plant height. Nodes (except \textit{Salix herbacea}, which represents the vegetation on the northern face of the gradient) are placed from left to right following an ascending order of their response to Snow duration (regression coefficient from the Generalized Linear Model used as a Habitat Suitability Model).  } \label{assocplant:b}
	\end{subfigure}
	\newline
	\centering
	\caption{Plant associations on an Alpine mesotopographic gradient. We highlight the communities (node colors) in figure (b) using colored labels on the matrix (a). } \label{assocplant}
	\end{figure}

The application of our approach to the Alpine dataset identified four densely connected modules of different sizes, within which species occupied distinct structural roles in the plant association network. Modules were structured along a gradient of response to the snow melting date (Fig.~\ref{assocplant}). 

\noindent Species from early-melting sites clustered into the same module, characterized by a dominance of positive associations—notably, a largely asymmetric attraction of forbs and grasses toward tall, dominant graminoids such as Carex and Kobresia. In contrast, forbs and grasses also formed two distinct groups connected by negative associations, indicative of competitive exclusion. Some of these species acted as hubs, linking high-elevation sites to adjacent zones where they also occurred.

\noindent The second module encompassed two groups of grasses: tall herbs occurring in favorable conditions, which were primarily structured by negative associations reflecting amensalism and competition; and short herb meadows, exposed to zoogenic disturbance, which exhibited increased abundance when co-occurring with tall herbs, suggesting a facilitative interaction.

\noindent The third module represented chionophilous (cold-adapted) vegetation found on late-melting sites. The fourth module encompassed north-facing, isolated communities, dominated by Salix herbacea, which showed positive associations with high-altitude communities but remained disconnected from other modules (Fig.~\ref{assocplant}).

\noindent Interestingly, we found a higher proportion of positive associations in communities from stressful environments, such as early-melting sites exposed to wind and erosion due to snowmelt \citep{choler2005consistent}. These associations likely reflect facilitative interactions mediated by graminoids through several mechanisms: graminoids help stabilize the soil \citep{callaway2007positive, heilbronn1984plant}, reduce desiccation and frost heaving on stones—thereby supporting seedling survival \citep{choler2001facilitation}—and create favorable microclimatic conditions that shelter smaller forbs and grasses from wind exposure \citep{wardle1998can}.
In contrast, negative associations were more frequent in species-rich sites, likely driven by competition for limiting resources such as water and nitrogen \citep{choler2001facilitation}. 


\newpage
\section{Discussion}
In this work, we tackled the challenge of inferring interspecific associations from multiple species co-abundances in hetereogeneous environments. To do so, we formalized pairwise associations as a function of two sets of latent variables representing the response and the effect of each species with respect to the others. We incorporated these associations into a conditional probabilistic model of abundance that accounts for environmental covariates. We evaluated our approach on both simulated and empirical datasets.

\subsection{Disentangling abiotic and biotic drivers}
\subsubsection{Uncovering positive, negative and neutral associations}
Comparatively to other tested frameworks, our method (EA) performed well on both positive and negative associations detection, despite the constraint induced by the embedding-based factorization. On average, discriminating positive and negative associations was within reach of most methods, provided an appropriate pair of thresholds was used to delimit the range of neutral associations. An exception concerned HMSC, which recovered multiple spurious associations that were potentially indirect effects despite the high support level. A more appropriate approach would have been to analyze the inverse covariance matrix. However, inverting the posterior estimate of the covariance matrix suffered from various numerical instabilities. 

Due to the upper-limit constraint of the fixed carrying capacity on the total count, all models inferred spurious negative associations between non-interacting species, especially in simulations with positive effects only, as a compensation mechanism. Association strengths were sensitive to niche overlap. For all methods, positive associations were easier to detect between species with overlapping niches. The fact that this pattern was observed for all methods as well as on the pairwise relative abundance indices suggested that the abiotic filter outruled these associations during the community assembly simulation.  

Amongst the tested methods, HMSC, PLN/EMtree and EcoCopula first fit the abiotic response then species dependencies are estimated either as random effects (HMSC, PLN) or from the marginal residuals (EcoCopula). The implicit importance given by these inference procedures to the abiotic drivers over the species associations explains the low detection rate of negative associations. In constrast, EA and MRFcov which both rely on an explicit regression over species abundances, do not suffer from the same bias, explaining their superiority in detecting negative effects.

\subsubsection{Uncovering prey-predator associations}
Several studies discuss the difficulties of recovering biotic interactions from co-occurrences \cite{sander2017ecological, barner2018fundamental, blanchet2020co}. In our experiment, we assumed that trophic interactions induce a dependence of predators on their preys but not vice-versa (directed positive association). 

We showed that using an appropriate coupling with the abiotic drivers allows to detect such associations providing that the species pair co-occur. However, the model detected symmetric dependencies when the abiotic niches of the pair overlapped strongly and especially in trophic chains and when the predator did not have other alternative prey. Moreover, the varying performances in recovering the true network structure for different food web topologies questions the power of the response-effect factorization to represent arbitrary directed acyclic graph (DAG) structures and suggests that a symmetric approach might be more effective, if coupled with knowledge of species trophic levels.  

\subsubsection{Importance of the abiotic-biotic aggregation function}
Species joint responses to abiotic (environment) and biotic (associations) drivers take on different forms, modeled by an aggregation function. Most existing frameworks are limited to linear or additive forms. Linear responses are particularly useful when associations are mediated by the environment (e.g in competition) or can alter it (as in habitat facilitation). In this case, associations compensate the suitability of the environment by either improving micro-habitat conditions or exerting a negative force that counterbalances it. 

On the other hand, when associations arise from direct interferences, their detection requires conditioning on co-occurrence, hence on habitat suitability \cite{gravel2019bringing}. When we fitted an additive architecture to the predator-prey occurrences, the model had a very low detection rate confirming that linear combinations of habitat suitability and biotic effects are not sensitive to such direct associations. These results may be specific to presence/absences and not hold true for abundances.

In general, the choice of an aggregation function depends on the type of interactions expected in the studied system. To guide this choice, several frameworks \citep{kissling2012towards, boulangeat2012accounting,thuiller2013road} conceptualize the incorporation of eco-evolutionary processes into species distribution models (a.k.a \textit{BI-SDMs} \citep{dormann2018biotic}). Besides, theoretical developments extended the theory of island biogeography \citep{macarthur2001theory} to account for trophic interactions \citep{gravel2011trophic} and more general interaction networks under environmental constraints \citep{cazelles2016theory, cazelles2016integration}.

\subsection{From species representations to biotic associations}
\subsubsection{The meaning of species embeddings}
In theory, the effect embedding of a species is equivalent to a factor analysis of all other species abundances (residual abiotic responses if coupled with environmental data) when it is present. The effect embedding is a proxy of the species' influence on the community composition. Combining the effect embeddings of occurring species produces an ordination of the community composition in the embedding space of dimension d: $R^d$. The species response embedding can be mapped into the same space, we can measure through the dot product the compatibility of the species to the observed community.

Since the community ordination is obtained as a linear combination of present species' effects, species response to the community can be rewritten as a sum of one-to-one responses to each observed species. When the response and effect embeddings are forced to be similar, we recover the same structure used by Latent Variable JSDMs. 

Analogously to the species embeddings, \cite{kissling2012towards} proposed the concept of \textit{interaction currencies} as surrogates for biotic interactions in distribution models in a similar response-effect framework. Hypothetically, these currencies include resources, bionomic variables \citep{hutchinson1957multivariate}, traits, and other non-consumable environmental conditions that mediate interactions. Our analysis of embeddings learnt from data in food web simulations showed that they captured both abiotic and biotic species preferences. In the case study on Alpine plants, we found that embeddings were mildly related to functional traits.   

\subsubsection{Constraining embeddings with prior knowledge}
In practice, the embedding dimension is typically significantly smaller than the number of species. While species can have distinct habitat preferences, the biotic role expressed in their interactions and the spatial associations they produce is drawn from a limited number (significantly smaller than the pool size) of behaviors represented by functional groups \citep{walker1992biodiversity}. A species can belong to one or several functional groups with different proportions. Such information can be mined from online databases or provided by experts \citep{betsi2012database,nguyen2016funguild,kattge2020try}. While learning graphical models with large species pools requires large datasets, replacing species with fixed groups \cite{ohlmann2018mapping} has two advantages: (1) to reduce the parameter space size by sharing embeddings within groups, (2) allowing extrapolation to new settings where different taxa are observed yet from the same modeled groups.  Besides, as evidenced by our simulated experiment, using group constraints can improve the ability of inference models to recover potential associations even when species did not co-occur. 

\subsection{Perspectives}
Beyond group constraints, some frameworks \citep{lo2017mplasso,chiquet2018variational, scutari2019package} support white-lists and black-lists, containing authorized and forbidden associations respectively, by penalizing graphs that do not satisfy those constraints. When interaction networks can be described at least partially, the same approach can be used to complete missing edges by harnessing similarities of species interactions. This semi-supervised problem is referred to as \textit{collaborative fitlering} \citep{fu2019link} and is one of the main applications of dependency networks. Incorporating this link prediction task within a multispecies distribution model would allow to quantify the effect of known and predicted interactions on species distributions.

We motivated throughout our simulation experiments the use of different joint responses for abiotic and biotic drivers depending on the underlying biotic interactions. The fact that interactions require and affect co-occurrences simultaneously are not mutually exclusive \citep{gravel2019bringing}. The availability of multi-trophic communities datasets \citep{derocles2018biomonitoring} where complex interactions are entangled calls for applications coupling different modes of aggregating abiotic drivers with biotic associations. 

\section{Conclusion}
Biological interactions and other processes induce spatial patterns of co-occurrence and co-abundance. We presented and validated a model of species co-abundances as a function of the habitat and biotic associations. We proposed an asymmetric scheme for modeling associations that is based on learning latent representations of species' responses and effects. Future efforts should be directed towards an incorporation of prior knowledge of the complete or partial topology of the association networks to guide the inference process. Along with that, a strong theory of how known ecological interactions influence the co-distribution of species is needed to support all these models.

\newpage

\bibliographystyle{plainnat}


\end{document}


\date{}
\maketitle
\tableofcontents

\newpage
\section{Supplements to framework description}
\subsection{Extensions of the biotic context definition}
\label{sup:alt-bio}
\subsubsection{Adding conditioning covariates}
In the base model, the estimation of any pairwise association is oblivious to the abiotic or biotic conditions surrounding it. To account for these neighborhood conditions, we extend the base model by allowing the embeddings used to represent the biotic context to vary according to some covariates.

Each site is associated to $p$ conditioning covariates. These covariates are stored alongside an offset in a $n \times (p+1)$ matrix $V$, such that each of the first $p$ columns of $V$ contains the values of the corresponding covariate for the different sites while the last column is filled with ones. 
Then, given an embedding dimension $d$, the covariates are mapped to $d$ dimensions by applying a regression with a weight matrix $W \in \mathbb{R}^{(p+1) \times (d)}$. The resulting conditioning vectors $\condv_\site = Wv^{T}_{\site}$ represent the relative weight associated to a given latent dimension depending on the conditions defined by the covariates.\\

The extended biotic context is then written as follows, where $\odot$ is the element-wise vector product: 
\begin{equation*}
  \BioEff_{\site\spc} = \condv_{\site} \odot \big(\frac{1}{\abs{\BioCont_{\site\spc}}} \sum_{\ospc \in \BioCont_{\site\spc}} \abdv_{\site\ospc} \Effv_{\ospc} \big)
  = \frac{1}{\abs{\BioCont_{\site\spc}}} \sum_{\ospc \in \BioCont_{\site\spc}} \abdv_{\site\ospc} \cdot (\condv_{\site} \odot \Effv_{\ospc}) 
\end{equation*}

The biotic associations can be recovered as in the base model, by isolating the pairwise associations in the response variable. However, in this case, the associations we obtain are represented by a three-dimensional tensor instead of a two-dimensional matrix. Each slice along the first dimension of this tensor represents a local association network.
\begin{align*}
  \Assv_{\site\spc\ospc} &= \sum_{l=1}^{d} (\condv_{\site} \odot \Resv_{\spc} \odot \Effv_{\ospc})_l \\
  \eta_{\site\spc} &= \linkf\big( \sum_{\ospc \in \BioCont_{\site\spc}} \abdv_{\site\spc}  \Assv_{\site\spc\ospc} + o_j\big)
\end{align*}

By incorporating the association covariates on the latent space, we gain two desirable properties. First, we get a fixed number of parameters that is a factor of the embedding dimension, which is significantly smaller than the number of modeled species. Second, we ensure species with similar latent traits, as captured by the response and effect embeddings, share associations regardless of the surrounding conditions. 

\subsubsection{Temporal extension}
When longitudinal data are available, we denote the abundance of species $\spc$ at site $\site$ at time-point $t$ as $\abdv_{\site\spc}^{(t)}$. Accordingly, the definition of the biotic context for a target species at a given time-point is extended to contain the species, including the target, that were observed in the previous time-point:
\begin{align*}
  \BioCont_{\site\spc}^{(t)} &= \{\ospc \in \spcS, \abdv_{\site\ospc}^{(t-1)} > 0 \} \\
  \BioEff_{\site\spc}^{(t)} &= \frac{1}{\abs{\BioCont_{\site\spc}^{(t)}}} \sum_{\ospc \in \BioCont_{\site\spc}^{(t)}} \abdv_{\site\ospc}^{(t-1)}  \Effv_{\ospc}
\end{align*}

\subsubsection{Spatial extension}
Given a function $\dst$ that measures the distance between any pair of sites and a radius $r$, we consider a spatial extension of the base model where the biotic context is defined to contain species that were observed at locations within distance $r$ of the considered site.
\begin{equation*}
  \BioCont_{\site\spc} = \{(\ospc, l) \in \spcS \times \siteS, \abdv_{l\ospc} > 0 \text{ and }  \dst(\site, l) \leq r\}
\end{equation*}

One can adapt the radius values to each species or group of species. The effect of each contextual element decays with distance to the target location, with a rate $\tau$ which controls the decrease in weight per unit of distance. 

\begin{equation*}
  \BioEff_{\site\spc} = \sum_{(\ospc, l) \in \BioCont_{\site\spc}} \abdv_{l\ospc} \cdot \exp(-\tau \dst(\site, l))
\end{equation*}

\subsubsection{Graph extension}
So far, we have defined the biotic context based on the local community composition, using the presence or abundance of other species to model pairwise effects on the target species. However, this formulation does not capture higher-order associations among the context species themselves, nor the broader network structure surrounding the target, what we refer to here as the \emph{contextual network}. To address this, we propose leveraging graph embedding techniques to compute latent representations of the contextual network.

\newpage
\section{Supplements to the virtual experiment 1}
\subsection{Simulation how-to}
\begin{figure}[bthp]
	\includegraphics[scale=1]{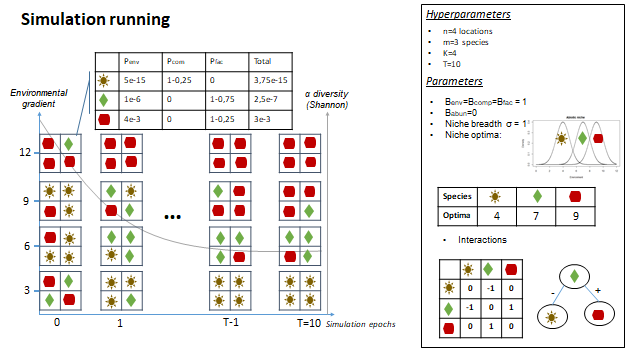}
	\caption{Simulation procedure.} \label{simillust}
\end{figure}

We used a process-based stochastic model adapted from Virtualcomm (Gallien and Münkemüller 2015) to simulate the assembly of individuals from a regional species pool into communities, on different locations sampled along an environmental gradient. The assembly process is controlled by three filtering mechanisms: the response to the abiotic environment, the outcome of biotic interactions and reproduction.  For simplicity, the spatial structure of communities and thus dispersal processes are ignored. In other words, there is no exchange of individuals between neighboring communities. The simulation starts with a given or random initial composition for each community independently. Individuals are replaced through time until an equilibrium state is reached or a user-defined number of iterations is completed. The final communities’ composition is returned at the end Fig.~\ref{simillust}.

\subsubsection{Notation}
\begin{itemize}
	\item[-] We start by sampling $n$ locations uniformly on a single environmental gradient $E$. 
	\item[-] All locations have the same carrying capacity of $K$ individuals from a common pool of $m$ species $S=\{S_j / j \in [1,m] \}$.
	\item[-] Each species has its own optimal environmental value $\mu_j \in E$ as well as a niche breadth $\delta_j \in E$. 
	\item[-] Biotic interactions are described by a full interaction matrix $I=(I_{jk}) / j,k \in [1,m]^2$ ; $-1 \leq I_{jk} \leq 1$ where $I_{jk}$ represents the effect of the interaction between the pair $(S_j, S_k)$ on species $S_k$. We also write: $I = I^+ - I^-  $ such that: 
	\begin{itemize}
		\item $I^+ =  (I_{jk}^+) / j,k \in [1,m]^2 ; 0 \leq I_{jk}^+ \leq 1 $ represents the matrix of positive effects (facilitation matrix)
		\item $I^- =  (I_{jk}^+) / j,k \in [1,m]^2 ; -1 \leq I_{jk}^+ \leq 0 $ represents the matrix of negative effects (competition matrix)
	\end{itemize}
\end{itemize}

\subsubsection{Assembly rules}
At each timestep (epoch), given an actual composition $c$, the probability that an individual from a given species $i$ to replace any other individual of $c$ is given by the following equation ; such that:
\begin{itemize}
	\item $B_{env}$: weights of the abiotic filter. 
	\item $B_{comp}$: weight of the competition.
    \item $B_{fac}$: weight of the facilitation.
	\item $B_{abun}$: weight of the reproduction filter, can be interpreted in terms of growth rate.
	\item $P_{env,i,c}$: the probability of species $i$ to occur under the environmental value $E_c$  is given by the normalized density on $E_c$ of a Gaussian distribution parameterized by its optimum and niche breadth. The closer to its optima, the higher the probability of the species’ occurrence.   
	\item $P_{comp,i,c}$: the probability for an individual of species $i$ to join the community given the aggregated effect of its competitors in $c$.  
	\item $P_{fac,i,c}$: the probability for an individual of species $i$ to join the community given the aggregated effect of its facilitators in $c$. 
	\item $P_{abund,i,c}$: probability of an individual of species $i$ to join the community as a result of the reproduction of some of the $N_{i,c}$ conspecifics in $c$.  
\end{itemize}

The unnormalized weights $W_{i,c}$ for each species are then normalized by dividing each one of them by their sum. The result is a vector of probabilities W that sums to 1. Finally, we sample from a multinomial distribution, parameterized with $W$, $K$ individuals to compose the new community.

\subsection{Simulation experiment}
\begin{figure}[H]
	\includegraphics[scale=0.75,angle=90]{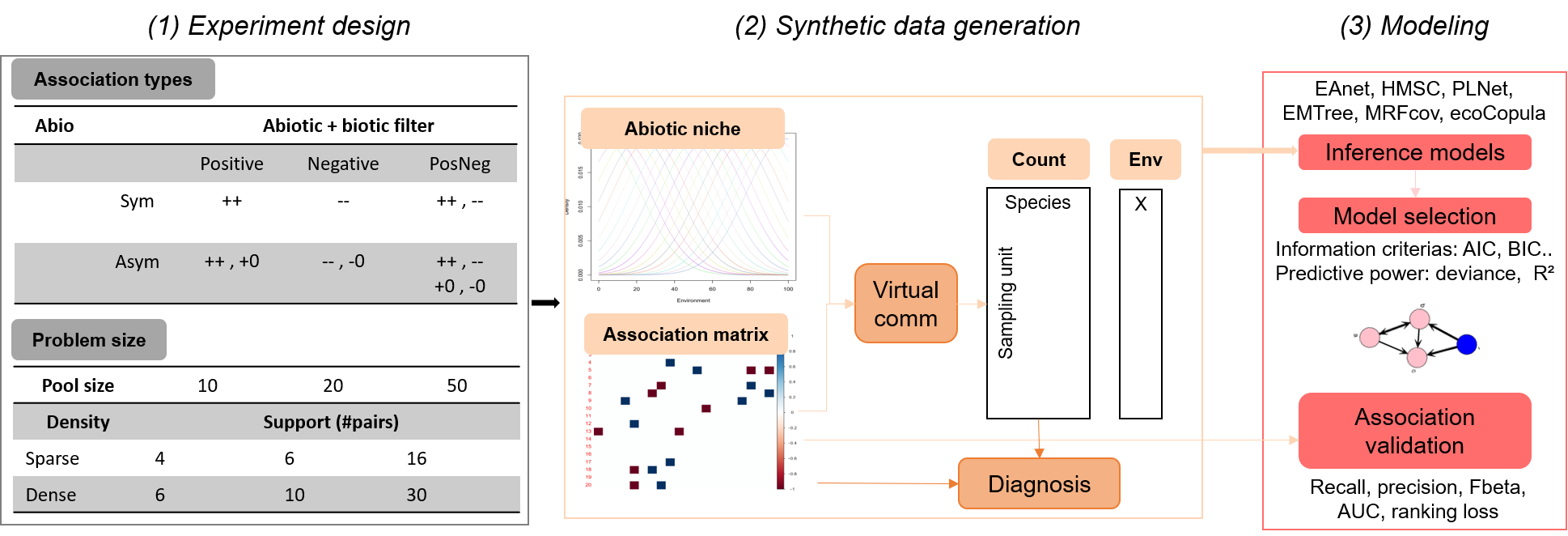} 	
	\caption{Description of the simulation experiment 1 with different assembly rules} \label{simexp1}
\end{figure}

We set up an experiment where multiple simulations were run on random locations along a single environmental gradient ranging from 0 to 100 with different randomly selected configurations of the prior association matrix: absence of association (environmental filtering only), positive associations only, negative associations only and a mix of positive and negative associations. In each configuration mode, we varied the pool size, i.e. the number of species (10, 20 or 50),  to test how the different models might be affected by the species pool size, the \textit{density} in terms of \textit{number of associated pairs} as a function of the pool size (sparse $1/3$ or dense $2/3$) and whether the association matrix included asymmetric effects: semi-attraction (e.g. commensalism) or semi-repulsion (e.g. amensalism). Positive (resp.\ negative) effects were all set to $+1$ (resp.\ $-1$) as we are interested in the polarity of the associations rather than their intensity. The factorial design of this experiment produced 33 simulation datasets ~Fig \ref{simexp1}. These combinations allowed us to test our framework, but also to compare its ability to detect species symmetric associations in respect to other approaches like JSDMs and probabilistic graphical models. 

\subsection{Simulation diagnosis}
To assess whether the simulated virtual communities reflect the simulation parameters (e.g. two competing species tend not to co-occur), we defined the \emph{relative abundance index} ($\text{RAI}_{ji}$), an asymmetric pairwise index that measures the average change in abundance of the target species $i$ when the source species $j$ is present as compared to its mean abundance irrespective of whether the source species $j$ is present, $\bar{\abdv}_{i}$. 

\begin{align*}
\Delta_{\site ji}  &= \{\, \abdv_{\site i} - \bar{\abdv}_{i},\text{ for all }\site \in \siteS\text{ such that }\abdv_{\site i} > 0 \text{ and } \abdv_{\site j} > 0\,\}\,.
\end{align*}

Then $\text{RAI}_{ji} = \avg(\Delta_{\site ji})$ across all $k$ sites.
The larger the standard deviation $\std(\Delta_{\site ji})$, the more ambiguous the strength of the effect of species $j$ on species $i$. If the confidence interval $\avg(\Delta_{ji}) \pm 1.96 \std(\Delta_{ji})$ does not contain zero, then the simulated dependencies unambiguously translate a polarized effect of species $j$ on species $i$. Otherwise, the simulations led to an equilibrium for which it is not possible to retrieve the parameters. 



\begin{figure}[bthp]
	\includegraphics[scale=1]{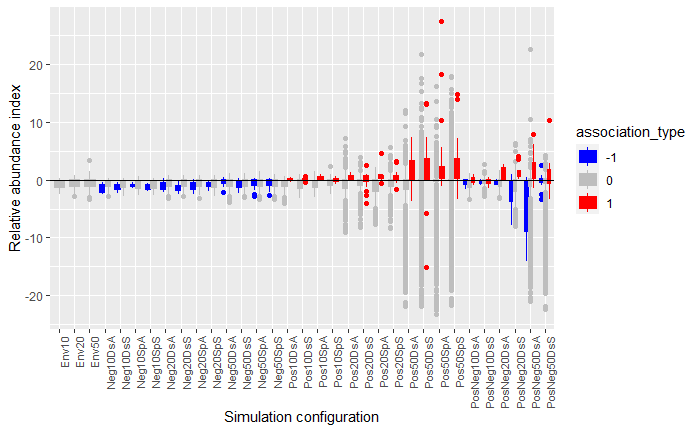} 	
	\caption{\textbf{Simulation diagnosis}. Distribution of relative abundance indices \evRAI{} per simulation. Each data point represents a directed association (positive in red, negative in blue and neutral in gray) involving two species from the corresponding simulation. Labels on the x-axis correspond to simulation configurations.} \label{sup:vc1patterns}
\end{figure}

Before fitting the inference models on the simulated data, we checked using an empirical measure of pairwise association \evRAI{} whether species dependencies reflected the simulated patterns. Fig~\ref{sup:vc1patterns} depicts the value of the statistic for each directed association $a_ij$. The distribution of \evRAI{} values showed a good discrimination of positive and negative associations, albeit with different strengths. Neutral associations translated into small \evRAI{} values, median-centered on zero for simulations with negative associations only. Whereas they spanned a large spectrum of values in simulations with only positive or a mix of positive and negative effects. The proposed indicator is itself a good proxy for association inference, however since it does not account for environmental covariates we use it as a diagnosis tool. 

\subsection{Supplementary results}
\subsubsection{Sensitivity of associations to niche distances}

\begin{figure}[H]
	\centering
	\includegraphics[scale=.75]{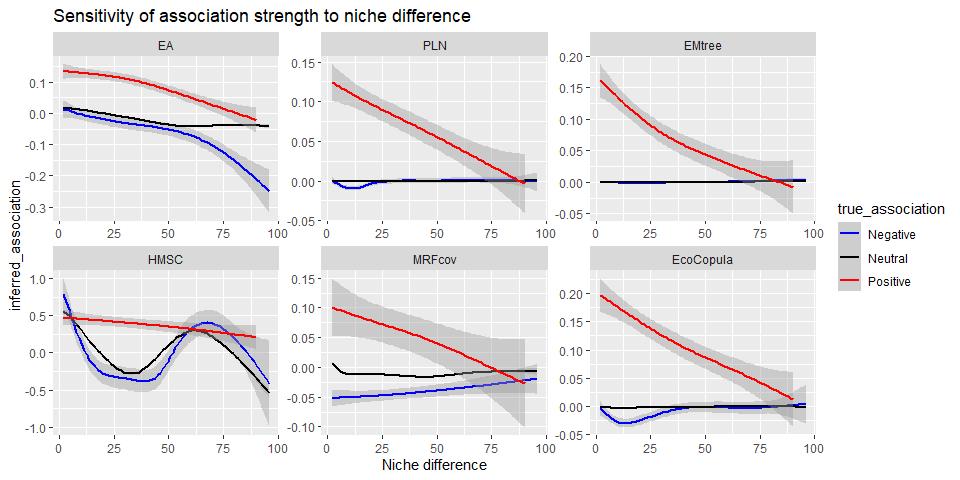} 	
	\caption{Sensitivity of the inferred association strength $a_{ij}$ per association type and inference model to the abiotic niche distance measured by the absolute difference between their niche optima $\mu_i - \mu_j$.} \label{exp1nichediff}
\end{figure}

On larger pool sizes, the distance between niche optima decreases as the probability of niche overlap increases. Consequently, the estimates of species associations varied with the number of modeled species. Although the strengths appeared to be drawn from a small fixed interval, their value was sensitive to the species niche differences (Fig ~\ref{exp1nichediff}). For all tested frameworks, the strength inferred for true positive associations decreased with the amount of niche overlap. For large niche differences (no overlap), MRFcov, EMtree and PLN even reported opposite signs. Conversely, inferred strengths of true negative associations were either invariant to niche difference (MRFcov), very small and close to neutral (EMtree, PLN, ecocopula and EA) or increasing in absolute strength (HMSC). At medium to high niche distance, EA reported an increase in the absolute strength of negative associations. HMSC showed the opposite pattern, suggesting that the negative effects were rather explained by the abiotic covariates.

\subsubsection{Comparative performances on association type inference}

\begin{figure}[bthp]
	\includegraphics[scale=.7]{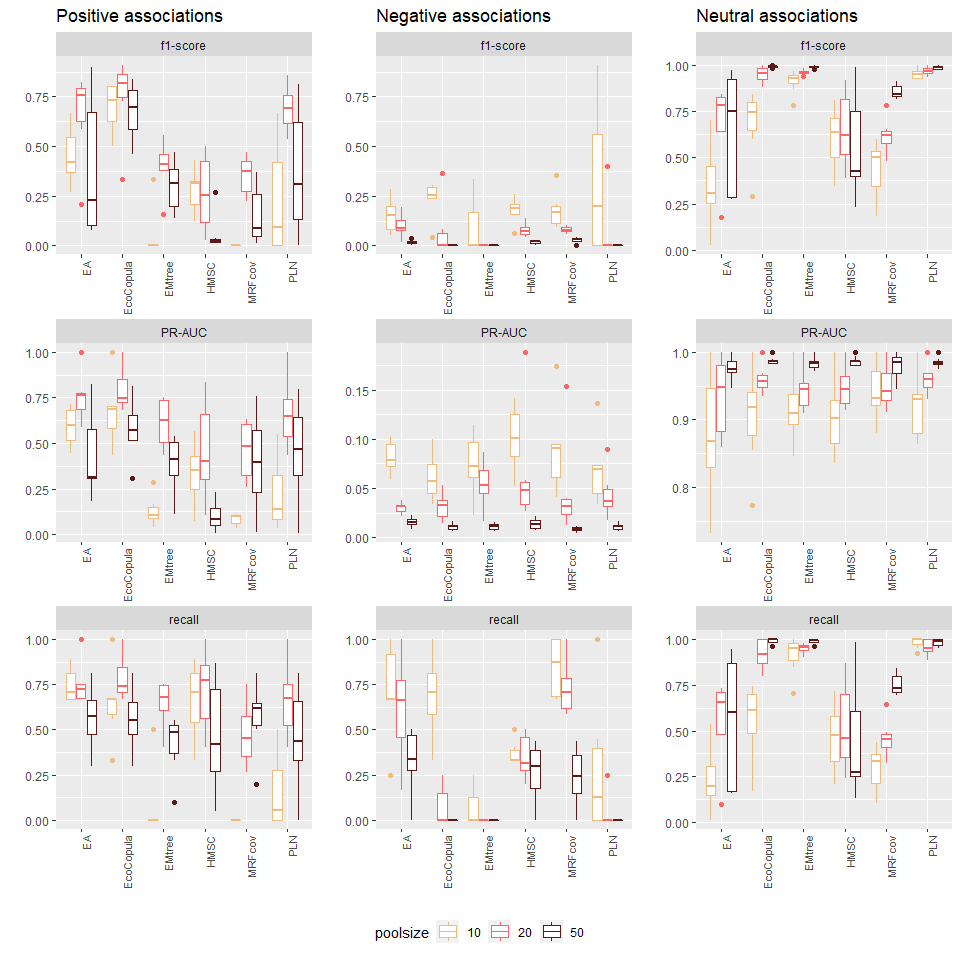} 	
	\caption{Inference of true association class per type of association for each model measured by the recall, f1-score and AUC-PR metrics. The higher the value the better performing is the model. AUC-PR is computed on the raw associations while the recall and f1-score are computed on the discretized associations using as a threshold $\epsPos=\epsNeg=5E-2$ 
	} \label{exp1perfs}
\end{figure}

We reported the Area Under the Precision-Recall Curve (PR-AUC), the recall and f1-score in Fig~\ref{exp1perfs} for each association type separately. We found no significant difference between the models in respect to their performances in inferring positive vs dense or symmetric/asymmetric datasets. However, the quality of inferred associations varied with the pool size. 
\\
\noindent On positive associations, EA and Ecocopula outperformed the other methods in all pool sizes. EMtree, PLN and MRFcov reported good performances for 20 and 50 species datasets, but they failed to detect positive associations in 10-species datasets. On negative associations, all models had strong difficulties in retrieving them, and this difficulty was more pronounced for large pool size. EA, MRFcov and HMSC outperformed other methods.   
 
\newpage
\section{Virtual experiment 2}
In this experiment, we simulated species occurrence data where each species depends jointly on suitable environmental conditions and the presence of at least one prey given by a known predator-prey network (food web). This setup reflects an intersection of abiotic and biotic filters, modeled as a multiplicative response. This multiplicative structure and the asymmetry of predator–prey interactions is not supported by other association inference methods. Therefore, we only evaluated our approach.

\subsection{Simulation how-to}

\begin{figure}[bthp]
	\includegraphics[scale=0.5]{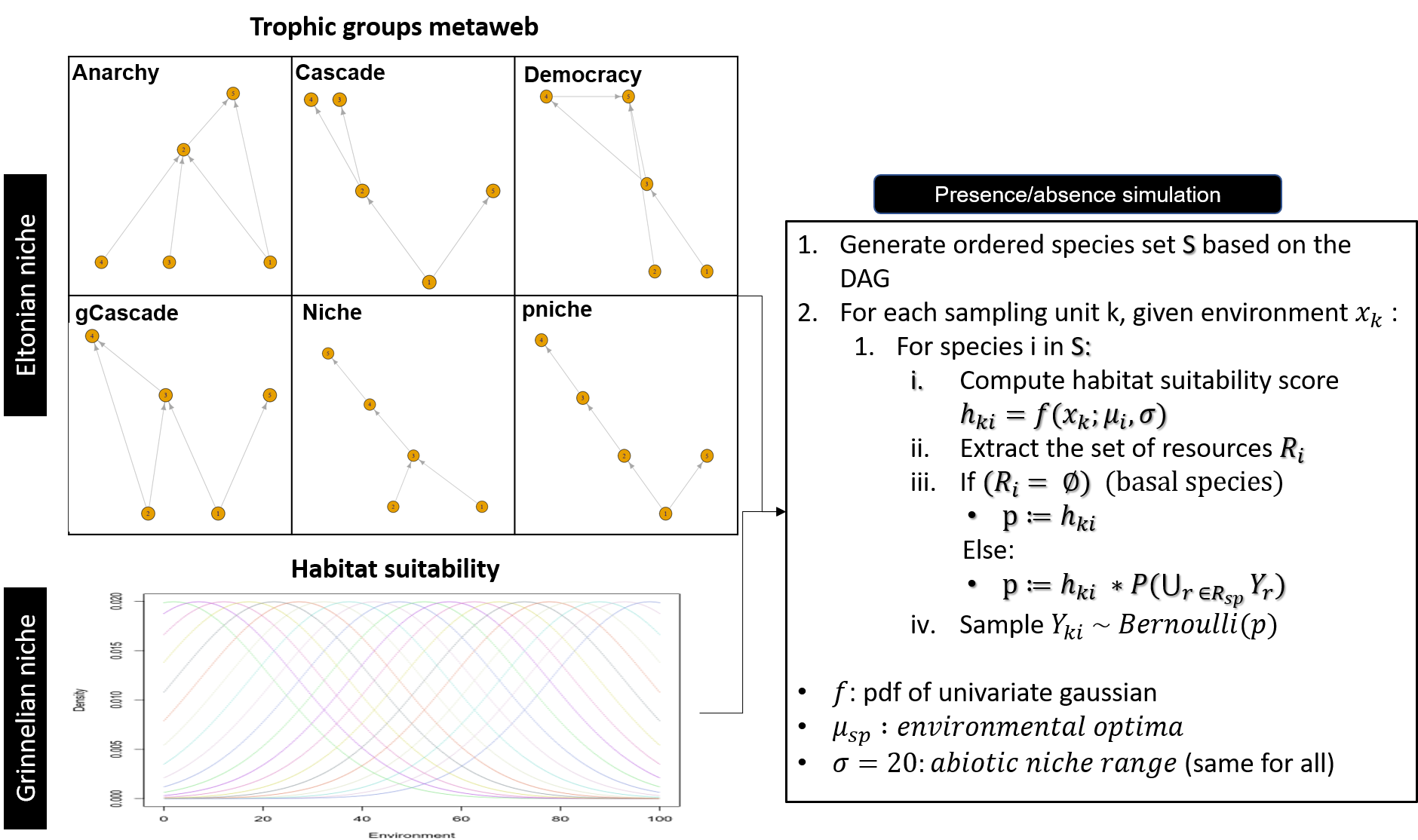} 	
	\caption{Simulations of consumer-resource co-occurrences along an environmental gradient given food web topology and true abiotic niches.} \label{simexp2}
\end{figure}

Fig~\ref{simexp2} illustrates the experimental setup and the procedure used to generate occurrences. Briefly, we assume a bottom-up control so that to be present, a consumer requires the availability of \textit{at least} one resource, in addition to habitat suitability. Basal trophic groups do not depend on any resource, only on the environment. The process resulted in six datasets, each containing 25 species and 500 sites.

\subsection{Structural regularities conserved in the embeddings}
For each topology, we investigated whether learnt embeddings reflect the underlying clustering of species into trophic groups. Concretely, we performed a Mann-Whitney ranking test \cite{mann1947test} checking whether species from the same prior trophic group had more similar embeddings than species from different groups. Finally, we asked to what extent response and effect embeddings captured species abiotic preferences and their biotic requirements. We did that by testing the correlation between embedding similarity and niche overlap measured with environmental optima differences and the proportion of shared preys in both the potential and realized food webs. 

\subsection{Supplementary results}

\subsubsection{Network structure inference performances}

The inferred associations were more faithful to the realized than the potential network. In all cases, incorporating a parameter-sharing constraint within trophic groups allowed to improve the sensitivity with respect to both ground truth networks. 

The main source of error was the confusion of directed associations with symmetric reciprocal associations, introducing spurious associations. This error was particularly frequent between species of inner groups in trophic chains.  However, the model managed to break this symmetry for species from groups with a higher diversity of preys (number of distinct groups) of preys. 

\begin{table}[bthp]
	\centering
	\resizebox{\textwidth}{!}{
\begin{tabular}{l|l|l|l|l|l|l|l|l|}
	\hline
	~ &\multicolumn{4}{|c|}{\textbf{Additive abiotic and biotic responses}} & \multicolumn{4}{c|}{\textbf{Intersection of abiotic and biotic requirements}} \\ 
	\hline
	~ &\multicolumn{2}{|c|}{\textbf{Group-level}} & \multicolumn{2}{c|}{\textbf{Species-level}} & \multicolumn{2}{c|}{\textbf{Group-level}} & \multicolumn{2}{c|}{\textbf{Species-level}} \\ 
	\hline
	~ & \textbf{MW}     & \textbf{RN}    & \textbf{MW}    & \textbf{RN}    & \textbf{MW}     & \textbf{RN}     & \textbf{MW}     & \textbf{RN}     \\ 
	\hline
	\bf{Accuracy (ACC)} & ~ & ~ & ~ & ~ & ~ & ~ & ~ & ~\\
	\hline
	~~ min & 0.45 & 0.80 & 0.72 & 0.82 & 0.45 & 0.62 & 0.24 & 0.38\\
	\hline
	~~ max & 0.83 & 0.94 & 0.83 & 0.94 & 0.88 & 0.96 & 0.72 & 0.87\\
	\hline
	~~ median & \textbf{0.790} & 0.890 & 0.765 & \textbf{0.895} & 0.780 & 0.735 & 0.615 & 0.735\\
	\hline
	~~ mean (sd) & 0.72 $\pm$ 0.15 & 0.87 $\pm$ 0.06 & 0.77 $\pm$ 0.04 & 0.88 $\pm$ 0.05 & 0.71 $\pm$ 0.18 & 0.76 $\pm$ 0.12 & 0.55 $\pm$ 0.19 & 0.69 $\pm$ 0.18\\
	\hline
	\bf{ROC-AUC (AUC)} & ~ & ~ & ~ & ~ & ~ & ~ & ~ & ~\\
	\hline
	~~ min & 0.24 & 0.07 & 0.49 & 0.50 & 0.50 & 0.75 & 0.47 & 0.62\\
	\hline
	~~ max & 0.80 & 0.85 & 0.53 & 0.74 & 0.80 & 0.92 & 0.58 & 0.83\\
	\hline
	~~ median & 0.470 & 0.415 & 0.505 & 0.565 & \textbf{0.670} & \textbf{0.855} & 0.535 & 0.800\\
	\hline
	~~ mean (sd) & 0.52 $\pm$ 0.22 & 0.43 $\pm$ 0.27 & 0.51 $\pm$ 0.01 & 0.58 $\pm$ 0.09 & 0.64 $\pm$ 0.12 & 0.84 $\pm$ 0.07 & 0.53 $\pm$ 0.04 & 0.77 $\pm$ 0.08\\
	\hline
	\bf{F2-score (F2)} & ~ & ~ & ~ & ~ & ~ & ~ & ~ & ~\\
	\hline
	~~ min & 0.00 & 0.00 & 0.00 & 0.00 & 0.29 & 0.50 & 0.27 & 0.49\\
	\hline
	~~ max & 0.56 & 0.65 & 0.19 & 0.44 & 0.68 & 0.77 & 0.52 & 0.55\\
	\hline
	~~ median & 0.120 & 0.220 & 0.105 & 0.220 & \textbf{0.520} & \textbf{0.590} & 0.415 & 0.520\\
	\hline
	~~ mean (sd) & 0.19 $\pm$ 0.23 & 0.25 $\pm$ 0.25 & 0.10 $\pm$ 0.08 & 0.19 $\pm$ 0.16 & 0.52 $\pm$ 0.14 & 0.61 $\pm$ 0.09 & 0.40 $\pm$ 0.11 & 0.52 $\pm$ 0.02\\
	\hline
	\bf{True Skill Statistic (TSS)} & ~ & ~ & ~ & ~ & ~ & ~ & ~ & ~\\
	\hline
	~~ min & -0.06 & 0.00 & -0.01 & 0.00 & 0.11 & 0.37 & 0.02 & 0.30\\
	\hline
	~~ max & 0.36 & 0.77 & 0.06 & 0.46 & 0.60 & 0.83 & 0.16 & 0.63\\
	\hline
	~~ median & 0.000 & 0.150 & 0.005 & 0.120 & \textbf{0.335} & \textbf{0.715} & 0.085 & 0.600\\
	\hline
	~~ mean (sd) & 0.08 $\pm$ 0.16 & 0.24 $\pm$ 0.30 & 0.02 $\pm$ 0.03 & 0.15 $\pm$ 0.17 & 0.34 $\pm$ 0.16 & 0.68 $\pm$ 0.16 & 0.09 $\pm$ 0.05 & 0.54 $\pm$ 0.13\\
	\hline
\end{tabular}
	}
	\caption{Summary of trophic network structure inference performances using two architectures: Additive abiotic and biotic effects, intersection of requirements (reported in the main text), with associations at the species level or the group level (by sharing embeddings within prior trophic groups). Metrics are computed taking in turn the potential \textbf{MW} (metaweb) and the realized \textbf{RN} network as ground truth. We highlight the best median score for each metric and reference network.}
	\label{tab:exp2perfs}
\end{table}

\subsubsection{Structural regularities captured by the embeddings}
For all topologies, the ranking test showed that both response and effect representations were significantly more similar when species belonged to the same trophic group. 
Moreover, species embedding similarity correlated positively with both abiotic and biotic niche similarity, with variable significance levels across topologies.  
Response and effect representations similarity were strongly correlated for two reasons: (1) species with similar responses were likely to have similar effects and vice-versa, (2) both representations were very similar, which explains the symmetry in some of the inferred associations.  

\begin{table}[bthp]
	\centering
	\resizebox{\textwidth}{!}{
\begin{tabular}{l|l|l|l|l|l|l}
	\hline
	& \textbf{Anarchy} & \textbf{Democracy} & \textbf{Cascade} & \textbf{gCascade} & \textbf{Niche} & \textbf{pNiche}\\
	\hline
	\bf{Internal clustering validation} & ~ & ~ & ~ & ~ & ~ & ~\\
	\hline
	~~ response & 38278(\textbf{<0.001}) & 36408(0.0021) & 40024(\textbf{<0.001}) & 35983(0.0044) & 37188(\textbf{<0.001}) & 38100(\textbf{<0.001})\\
	\hline
	~~ effect & 36683(0.0013) & 37300(\textbf{<0.001}) & 39164(\textbf{<0.001}) & 36708(0.0013) & 36448(0.002) & 38188(\textbf{<0.001})\\
	\hline
	\bf{Abiotic niche similarity} & ~ & ~ & ~ & ~ & ~ & ~\\
	\hline
	~~ response & 0.21(\textbf{<0.001}) & 0.13(0.0014) & 0.13(\textbf{<0.001}) & 0.16(\textbf{<0.001}) & 0.23(\textbf{<0.001}) & 0.16(\textbf{<0.001})\\
	\hline
	~~ effect & 0.21(\textbf{<0.001}) & 0.12(0.0033) & 0.18(\textbf{<0.001}) & 0.24(\textbf{<0.001}) & 0.07(0.065) & 0.15(\textbf{<0.001})\\
	\hline
	\bf{Eltonian niche similarity} & ~ & ~ & ~ & ~ & ~ & ~\\
	\hline
	~~ response & 0.04(0.29) & 0.1(0.015) & 0.15(\textbf{<0.001}) & 0.09(0.03) & 0.12(0.0039) & 0.13(0.0011)\\
	\hline
	~~ effect & 0.04(0.31) & 0.11(0.0052) & 0.2(\textbf{<0.001}) & 0.08(0.06) & 0.08(0.041) & 0.15(\textbf{<0.001})\\
	\hline
	\bf{Filtered prey similarity} & ~ & ~ & ~ & ~ & ~ & ~\\
	\hline
	~~ response & 0.14(\textbf{<0.001}) & -0.03(0.39) & 0.34(\textbf{<0.001}) & -0.11(0.0052) & 0.13(0.0013) & 0.03(0.48)\\
	\hline
	~~ effect & 0.13(\textbf{<0.001}) & 0.02(0.59) & 0.31(\textbf{<0.001}) & -0.09(0.033) & 0.05(0.25) & 0.08(0.035)\\
	\hline
\end{tabular}
}
\caption{Structural regularities captured by the response and effect embeddings, for each food web topology.  }\label{embstruct} 
\end{table}

\newpage
\section{Supplements to the empirical application}
\subsection{Environmental data preparation}
The plant dataset contained the following set of environmental variables:
\begin{description}
	\setlength\itemsep{-.2em}
	\item[\trtSlope]: the slope inclination in degrees,
	\item[\trtSnow]: the average snowmelt date in Julian days between 1997 and 1999,
	\item[\trtPhysD]: the percentage of non vegetated soil due to physical processes,
	\item[\trtZoogD]: the percentage of non vegetated soil due to marmot activity,
	\item[\trtAspect]: the relative south aspect, and
	\item[\trtForm]: the microtopographic landform index.
\end{description}

We initially applied a one-hot encoding scheme to the two categorical features (aspect and form) and we scaled the numerical features.

\subsection{Framework adaptation and training}
We split the observations into a training and a test dataset using a multi-label stratification scheme\footnote{Python library \texttt{scikit-multilearn}: \url{http://scikit.ml/}} to ensure that all species were covered and their proportions were preserved in both sets.

For each plant species, we pre-trained a generalized linear model (GLM) with a logit link to relate species occurrences to the environmental variables. We used the learnt weights as initial parameter values in the habitat suitability component of our framework.

We defined the biotic context for a target species as the set of plants observed on the location of interest. We used a negative binomial distribution to fit the plant counts. The embedding vectors were initialized using random samples from a uniform distribution on the $[-0.01,0.01]$ interval, and subjected to lasso penalties to promote sparsity. Finally, the offset value for each species was set to its average count on occurrence points.

We trained the full model using stochastic gradient descent (with a learning rate of $0.01$ and momentum of $0.8$) on the training dataset using a subsampling rate of $25\%$ for the negative examples. We monitored the negative log-likelihood of positive examples (presences) on the validation set after each full pass of the training set to assess the convergence of the training. We stopped when the loss stops decreasing or when $200$ epochs have elapsed.

\subsection{Embedding dimension and lasso parameter selection}
The first step in this evaluation was to find appropriate values for the hyperparameters of our model. For a species pool of size $m$, the embedding dimension $d$ is selected among powers of $2$ up to $m/2$, to improve hyperparameter search speed. In our case, with $m=82$, the embedding dimension is chosen from the set $\{2,4,8,16,32\}$.

When the value of the lasso penalty parameter $\lambda$ becomes large, some components of the embedding vectors take extremely small values for all species (below $10^{-5}$). These components have no effect on the computed associations. Removing them, shrinks the embeddings to a smaller effective dimension, equal to the number of retained components. In the extreme, very high values of $\lambda$ lead to effective dimension equal to zero, resulting in a zero association matrix, so that the interaction model is only parameterized by the species offset counts.

For each value of $d$, we apply the training procedure described previously with increasing values of $\lambda \in \{0.01,0.015,0.02,0.025\}$. We evaluate the resulting models on the test set by computing the effective dimension and the deviance of the predicted counts on positive examples (Fig.~\ref{gridaravo}).

\begin{table}[H]
	\centering
	\colorlet{cbckg}{TolDivA}

\begin{tabular}{cccccc}
  \toprule
  $\lambda$ \textbackslash $k$  & $2$ & $4$ & $8$ & $16$ & $32$ \\
 ~ \\[-.5em]
 \multicolumn{5}{l}{\quad Effective dimension} \\
\midrule
$0.010$ & \cellcolor{cbckg!15}$2$ & \cellcolor{cbckg!20}$4$ & \cellcolor{cbckg!30}$8$ & \cellcolor{cbckg!50}$16$ & \cellcolor{cbckg!90}$32$ \\
$0.020$ & \cellcolor{cbckg!15}$2$ & \cellcolor{cbckg!17}$3$ & \cellcolor{cbckg!22}$5$ & \cellcolor{cbckg!37}$11$ & \cellcolor{cbckg!62}$21$ \\
$0.030$ & \cellcolor{cbckg!10}$0$ & \cellcolor{cbckg!10}$0$ & \cellcolor{cbckg!12}$1$ & \cellcolor{cbckg!17}$3$ & \cellcolor{cbckg!10}$0$ \\
  $0.040$ & \cellcolor{cbckg!10}$0$ & \cellcolor{cbckg!10}$0$ & \cellcolor{cbckg!10}$0$ & \cellcolor{cbckg!10}$0$ & \cellcolor{cbckg!10}$0$ \\
\midrule
  ~ \\[-.5em]
\multicolumn{5}{l}{\quad Deviance} \\
\midrule
$0.010$ & \cellcolor{cbckg!13}$0.300$ & \cellcolor{cbckg!11}$0.295$ & \cellcolor{cbckg!12}$0.296$ & \cellcolor{cbckg!10}$0.290$ & \cellcolor{cbckg!10}$0.287$ \\
$0.020$ & \cellcolor{cbckg!14}$0.302$ & \cellcolor{cbckg!12}$0.298$ & \cellcolor{cbckg!12}$0.298$ & \cellcolor{cbckg!12}$0.295$ & \cellcolor{cbckg!12}$0.295$ \\
$0.030$ & \cellcolor{cbckg!90}$0.579$ & \cellcolor{cbckg!90}$0.579$ & \cellcolor{cbckg!14}$0.304$ & \cellcolor{cbckg!14}$0.305$ & \cellcolor{cbckg!90}$0.579$ \\
  $0.040$ & \cellcolor{cbckg!90}$0.579$ & \cellcolor{cbckg!90}$0.579$ & \cellcolor{cbckg!90}$0.579$ & \cellcolor{cbckg!90}$0.579$ & \cellcolor{cbckg!90}$0.579$ \\
  \midrule
  ~ \\[-.5em]
\multicolumn{5}{l}{\quad AIC} \\
\midrule
$0.010$ & \cellcolor{cbckg!10}$1148.960$ & \cellcolor{cbckg!15}$1804.961$ & \cellcolor{cbckg!26}$3116.962$ & \cellcolor{cbckg!47}$5740.960$ & \cellcolor{cbckg!90}$10988.957$ \\
$0.020$ & \cellcolor{cbckg!10}$1148.946$ & \cellcolor{cbckg!15}$1804.952$ & \cellcolor{cbckg!26}$3116.954$ & \cellcolor{cbckg!47}$5740.951$ & \cellcolor{cbckg!89}$10988.947$ \\
$0.030$ & \cellcolor{cbckg!10}$1148.704$ & \cellcolor{cbckg!15}$1804.704$ & \cellcolor{cbckg!26}$3116.936$ & \cellcolor{cbckg!47}$5740.920$ & \cellcolor{cbckg!89}$10988.704$ \\
  $0.040$ & \cellcolor{cbckg!10}$1148.704$ & \cellcolor{cbckg!15}$1804.704$ & \cellcolor{cbckg!25}$3116.704$ & \cellcolor{cbckg!47}$5740.704$ & \cellcolor{cbckg!89}$10988.704$ \\
  \bottomrule
\end{tabular}
	\caption{Effective dimension (number of non-zero components), positive deviance and Akaike Information Criterion (AIC) as a function of the embedding size (k) and the lasso penalty parameter $\lambda$.}
     \label{gridaravo}
\end{table}

\subsection{Habitat suitability}
\begin{figure}[H]
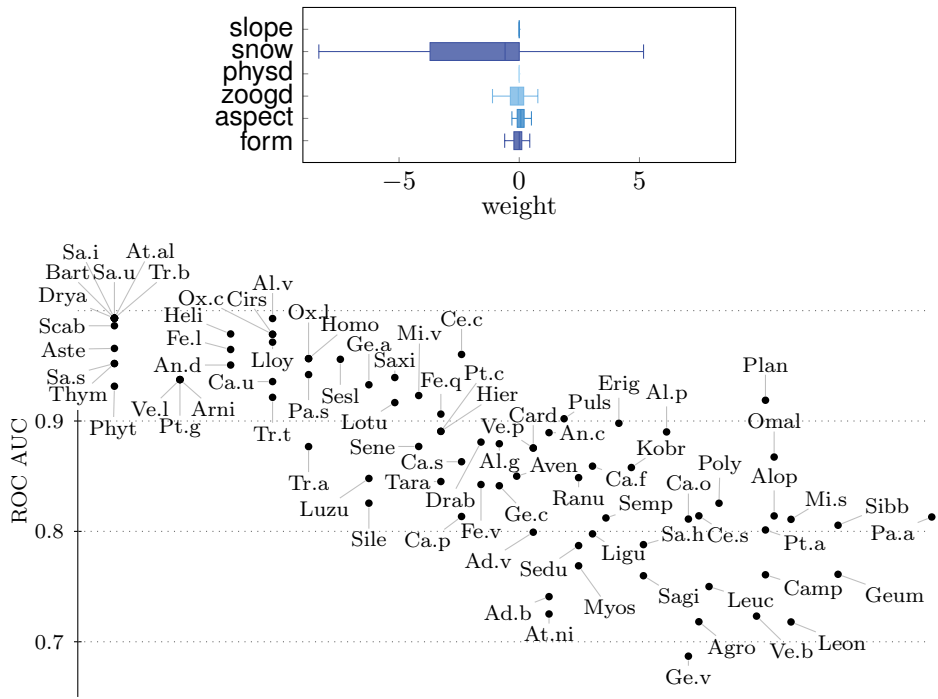

	\centering
	\inclPlt{7}{box_AlpsHSMw}
	\inclPlt{8}{scatter_AlpsPlantScores}
	\caption{Habitat Suitability Model variable importance and prediction performances per genus.}
	\label{hsmaravo}
\end{figure}

\paragraph{Performances}
The model predicts habitat suitability with a $87.7 \pm 0.17\%$ AUC score for all genera (Fig.~\ref{hsmaravo}). The analysis of environmental variable importance showed the dominance of snow duration followed by zoogenic disturbances, the site form and aspect. Physical disturbance and slope weights were negligible, probably due to their correlation with snow. 

\subsection{Summary association network}
We performed a hierarchical co-clustering on the inferred association matrix, to obtain effect and response groups. In parallel, we applied the modularity maximization algorithm \cite{newman2006modularity} on the association network to identify densely connected modules, referred to as communities (\textit{sensu} graph theory) \cite{gauzens2015trophic}. After that, we mapped the structural roles within the modules to create the group-level network. (See Supplementary Results). Finally, we analyzed the resulting patterns in light of existing literature on Alpine plants interactions \cite{choler2001facilitation}.

\begin{figure}[H]
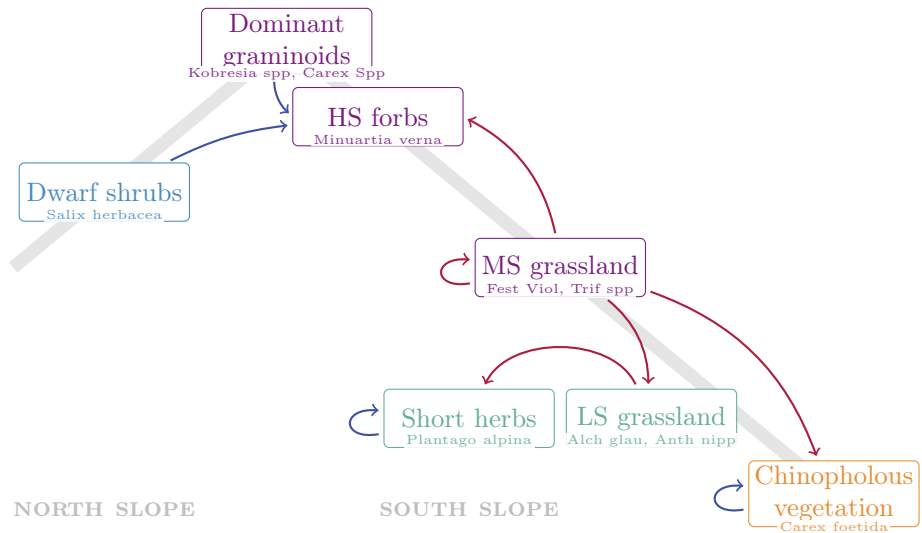

	\centering
	\inclPlt{11}{network_summary}
	\caption{The summary association network. Structural roles (nodes) are mapped to position in the gradient (Higher-slope HS, mid-slope MS, lower-slope LS) and plant classes (graminoids, grasses/herbs, forbs) and network modules (node colors). Edges go from a source (effect group) to a target (response group). Blue (resp.\ red) edges represent positive (resp.\ negative) associations.}\label{discussgroup}
\end{figure}

\subsection{Analyzing the functional meaning of plant embeddings}
We investigated the functional determinants of the associations diversity. To do so, we compute the mutual information between the learnt embeddings and the plant traits (reported in \cite{choler2005consistent}). The Mutual Information \cite{shannon1949mathematical} is an unbounded symmetric and positive score that measures the amount of information contained in one random variable about another. It quantifies the reduction in uncertainty about one random variable given knowledge of another. Zero mutual information indicates independence.\\

In general, we expect traits related to dispersal capabilities (seed mass, spread) to impact the prevalence of the species, consequently increasing or decreasing the opportunity to affect other species (interaction probability). As a result, we expect such traits to have a higher mutual information with effect embeddings than with response embeddings. Conversely, traits related to nutrient uptake and biomass accumulation potential capture competitive or cooperative abilities of the plant species. Hence, we would expect a high mutual information between these traits and both responses and effects embeddings.\\

\begin{figure}[H]
	\begin{subfigure}{\textwidth}
		\centering
		\includegraphics[page=9]{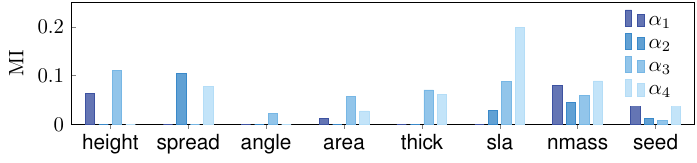}
	\end{subfigure}
	\newline
	\begin{subfigure}{\textwidth}
	    \centering
		\includegraphics[page=10]{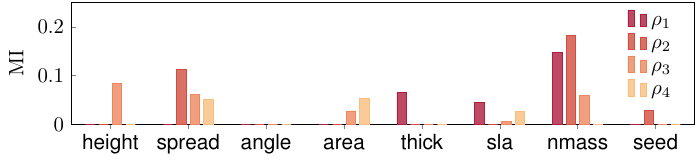}
	\end{subfigure}
	\caption{Mutual information between plant traits and their latent representations. Each bar concerns a specific trait, it represents the stack of mutual information scores from the first to the last (fourth) embedding dimension. The lower (resp. upper) figure shows the results for the response (resp. effect) embeddings.}\label{mitraitemb}
\end{figure}

There was a relatively significant contribution of the leaf nitrogen mass and spread to the plants response, whereas leaf angle was found independent (Fig.~\ref{mitraitemb}). The Specific Leaf Area contributes significantly to the effect in addition to the Nitrogen mass and on a lesser extent Spread. Height is reported as related to both parameters. \\

\bibliographystyle{acm}